\documentclass[journal]{IEEEtran}
\usepackage{amsmath}

\usepackage{cite}

\ifCLASSINFOpdf
\else
\fi
\usepackage[pdftex]{graphicx}          

\graphicspath{{./figures/}}

\DeclareGraphicsExtensions{.pdf}    
\usepackage{graphicx}
\usepackage{subcaption}
\usepackage{pgffor}

\usepackage[margin=1cm]{geometry}
\usepackage[percent]{overpic}   
\usepackage{subcaption}
\usepackage{xcolor}
\usepackage{pgffor}             

\usepackage{tikz}
\usetikzlibrary{calc}   
\usetikzlibrary{spy} 
\usepackage{booktabs}
\usepackage{multirow}
\usepackage{siunitx}
\usepackage{adjustbox}

\usepackage{url}

\begin{document}
%
\title{A TRPCA-Inspired Deep Unfolding Network for Hyperspectral Image Denoising via Thresholded
 t-SVD and Top-K Sparse Transformer
}
%
%
%

\author{Liang Li,
        Jianli Zhao,
        Sheng Fang,
        Siyu Chen
        and~Hui Sun
        
\thanks{Jianli Zhao is with the the College
 of Computer Science and Engineering, Shandong University of Science and Technology,
 Qingdao, China.\protect\\
E-mail: mailto:jlzhao@sdust.edu.cn.}
\thanks{Liang Li, Sheng Fang and Siyu Chen are with the College of Computer Science and Engineering, Shandong University of Science and Technology, Qingdao, China.}
\thanks{Hui Sun is with Gosci Information Technology Co.,Ltd., Qingdao, China.}

\thanks{Manuscript received April 19, 2005; revised September 17, 2014.}}

%
%

\markboth{Journal of \LaTeX\ Class Files,~Vol.~13, No.~9, September~2014}%
{Shell \MakeLowercase{\textit{et al.}}: Bare Demo of IEEEtran.cls for Journals}
%



\maketitle

\begin{abstract}
Hyperspectral images (HSIs) are often degraded by complex mixed noise during acquisition and transmission, making effective denoising essential for subsequent analysis. Recent hybrid approaches that bridge model-driven and data-driven paradigms have shown great promise. However, most of these approaches lack effective alternation between different priors or modules, resulting in loosely coupled regularization and insufficient exploitation of their complementary strengths. Inspired by tensor robust principal component analysis (TRPCA), we propose a novel deep unfolding network (\textbf{DU-TRPCA}) that enforces stage-wise alternation between two tightly integrated modules: low-rank and sparse. The low-rank module employs thresholded tensor singular value decomposition (t-SVD), providing a widely adopted convex surrogate for tensor low-rankness and has been demonstrated to effectively capture the global spatial-spectral structure of HSIs. The Top-K sparse transformer module adaptively imposes sparse constraints, directly matching the sparse regularization in TRPCA and enabling effective removal of localized outliers and complex noise. This tightly coupled architecture preserves the stage-wise alternation between low-rank approximation and sparse refinement inherent in TRPCA, while enhancing representational capacity through attention mechanisms. Extensive experiments on synthetic and real-world HSIs demonstrate that DU-TRPCA surpasses state-of-the-art methods under severe mixed noise, while offering interpretability benefits and stable denoising dynamics inspired by iterative optimization. Code is available at \url{https://github.com/liangli97/TRPCA-Deep-Unfolding-HSI-Denoising}.
\end{abstract}

\begin{IEEEkeywords}
Hyperspectral image denoising, deep unfolding, tensor robust principal component analysis, low-rank, sparse transformer.
\end{IEEEkeywords}

%
\IEEEpeerreviewmaketitle

\section{Introduction}
\IEEEPARstart{H}{yperspectral} images (HSIs) capture rich spectral information across numerous contiguous bands and have been widely used in remote sensing tasks such as object classification~\cite{zhang2018}, change detection~\cite{liuReviewChangeDetection2019}, and anomaly detection\cite{zhuDeepLearningRemote2017}. However, HSIs are often corrupted by various types of noise introduced during data acquisition and transmission, such as Gaussian noise, impulse (sparse) noise, and stripe noise\cite{rasti2018noise}. These degradations severely compromise image quality and hinder subsequent analysis. Effective HSI denoising is therefore a critical preprocessing step, yet it remains challenging due to the high dimensionality of HSI data and the complex mixtures of noise encountered in practice\cite{zhang2021hyperspectral}. 

Over the past decade, HSI denoising methods have evolved from purely model-driven approaches, through data-driven neural networks, and towards hybrid approaches that integrate both paradigms\cite{zhangHyperspectralImageDenoising2024}. Early model-driven approaches leveraged strong mathematical priors—such as low-rank assumptions\cite{chenFlexDLDDeepLowrank2024,changHyperLaplacianRegularizedUnidirectional2017,changWeightedLowRankTensor2020}, total variation (TV) regularization\cite{he2015total,heHyperspectralImageDenoising2018}, and nonlocal self-similarity\cite{he2020non,kong2020hyperspectral} to formulate HSI denoising as optimization problems. Among these, \emph{tensor robust principal component analysis} (TRPCA), implemented via tensor singular value decomposition (t‑SVD), explicitly decomposes an HSI tensor into low‑rank and sparse components, exhibiting strong robustness to mixed noise~\cite{luTensorRobustPrincipal2020}. However, despite their interpretability and empirical effectiveness, model-driven approaches typically require careful tuning of regularization parameters and can degrade under complex, real-world noise scenarios\cite{zhang2021hyperspectral}. 

To overcome these drawbacks, data-driven approaches have gained increasing attention by learning nonlinear mappings from noisy to clean HSIs, leveraging architectures such as convolutional neural networks (CNNs), recurrent neural networks (RNNs), and more recently transformer-based models. Representative models—including HSID-CNN\cite{yuan2018hyperspectral}, QRNN3D\cite{wei3DQuasirecurrentNeural2021}, and HSDT\cite{laiHybridSpectralDenoising2023}—have achieved impressive denoising performance by extracting rich spatial–spectral features. Transformers, with their global self-attention, are particularly effective for modeling long-range dependencies. However, despite these advances, purely data-driven models often require large amounts of paired training data, act as “black boxes,” and typically exhibit limited generalization and adaptability, especially when evaluated on different or more complex real-world data, due to the lack of explicit physical constraints\cite{zhangHyperspectralImageDenoising2024,xiongSMDSnetModelGuided2022}. 

To combine the strengths of both paradigms, hybrid approaches have emerged—integrating model-based priors with learning-based representations to enhance both interpretability and performance. However, the integration of multiple priors—such as low-rank and sparse structures—remains largely superficial. Plug-and-play (PnP) frameworks, such as Deep-PnP\cite{laiDeepPlugandplayPrior2022}, replace proximal operators with a fixed pretrained denoiser, whose weights are not updated or jointly optimized with the underlying data-fidelity model during inference. Meanwhile, architecture-guided designs (e.g., T3SC~\cite{bodrito2021trainable}, LR-Net~\cite{zhangLRNetLowRankSpatialSpectral2021}, MAC-Net~\cite{xiongMACNetModelAidedNonlocal2022}, HyLoRa~\cite{tanLowrankPromptguidedTransformer2024}) inject low-rank or sparse blocks into CNN/transformer backbones, but these hybrid models often fail to enforce strict stage-wise alternation between priors, limiting the mutual refinement. 

Deep unfolding networks (DUNs) offer a principled solution by mapping each iteration of an optimization algorithm onto a trainable network layer, theoretically enabling explicit and strict stage-wise alternation between different priors\cite{zhangISTAnetInterpretableOptimizationinspired2018,alhejailiRecursionsAreAll2023}. While DUNs have led to significant advances in natural image restoration, their application to high-dimensional HSI denoising remains limited\cite{guo2025deep,de2024deep,mouDeepGeneralizedUnfolding2022}. In practice, existing DUNs for HSI denoising are restricted to a few representative models, each with specific limitations: ILR-Net~\cite{yeIterativeLowrankNetwork2024} focuses solely on low-rank modeling; DNA-Net~\cite{zengDegradationnoiseawareDeepUnfolding2025} replaces explicit model-based priors with a transformer module, thereby introducing data-driven components that may not correspond to clear physical priors; and SMDS-Net~\cite{xiongSMDSnetModelGuided2022} combines a one-shot spectral projection with iterative sparse coding, lacking true alternation. None of these fully capture the complementary benefits or theoretical rigor of strict stage-wise alternation as established in TRPCA, thereby limiting their robustness to complex mixed noise. 

Motivated by the above analysis, we propose \textbf{DU-TRPCA}, a \textbf{D}eep \textbf{U}nfolding network inspired by \textbf{T}ensor \textbf{R}PCA for HSI denoising. Our network strictly enforces stage-wise alternation between two purpose-designed modules—a differentiable thresholded t-SVD for low-rank approximation, and a learnable Top-K sparse transformer for sparse refinement—at every iteration. Specifically, the low-rank module is motivated by the global spatial-spectral correlations inherently present in HSIs. The thresholded t-SVD serves to extract the principal low-rank components and preserve the intrinsic structure of the data, enabling effective reconstruction of the underlying clean HSI. Subsequently, the sparse module serves as a neural analog of the explicit sparse constraint in TRPCA. To embody this prior in our network, we employ a Top-K Sparse Transformer integrated into the deepest encoder layer of Hybrid Spectral Denoising Transformer (HSDT), which adaptively preserves only the most significant features and suppresses irrelevant responses. Crucially, the Top-K selection directly maps to the role of the sparse term in TRPCA—it enforces that only a minority of elements (i.e., features or spatial locations) can be highly activated, analogous to isolating sparse noise or outliers in the classic model. Recent works have demonstrated that Top-K sparse attention significantly enhances feature selection and robustness to structured noise in low-level vision tasks\cite{chen2023learning,zhou2024adapt}. Here, for the first time, we explicitly align this mechanism with the sparse regularizer in TRPCA, thus providing a principled, interpretable, and data-adaptive implementation of the sparse prior within a deep network. 

By alternating these two modules in a multi-stage unfolding architecture, DU-TRPCA inherits the interpretability and theoretical guarantees of TRPCA, while leveraging the powerful representational capacity and adaptability of modern transformers. This tightly coupled design enables mutual and iterative refinement between low-rank and sparse priors, leading to enhanced robustness and superior denoising performance.

The main contributions of this work are as follows:
\begin{itemize}
\item We propose a TRPCA-inspired deep unfolding network for HSI denoising that strictly enforces \emph{stage-wise alternation} between low-rank and sparse components, enabling tight structural coupling and iterative prior refinement.
\item We develop a \emph{differentiable thresholded t-SVD} module for tensor low-rank projection, effectively extracting the global spatial-spectral structure of HSIs.
\item We introduce a \emph{Top-K sparse block} integrated into the deepest encoder layer of HSDT, which explicitly enforces sparsity in the learned features, providing a deep neural realization of the TRPCA sparse prior.
\item Extensive experiments on synthetic and real noisy HSI benchmarks demonstrate that DU-TRPCA surpasses state-of-the-art methods under severe mixed noise, while offering interpretability benefits and stable denoising dynamics inspired by iterative optimization.
\end{itemize}

\begin{figure*}[!t]                      
  \centering
  \includegraphics[width=\textwidth]{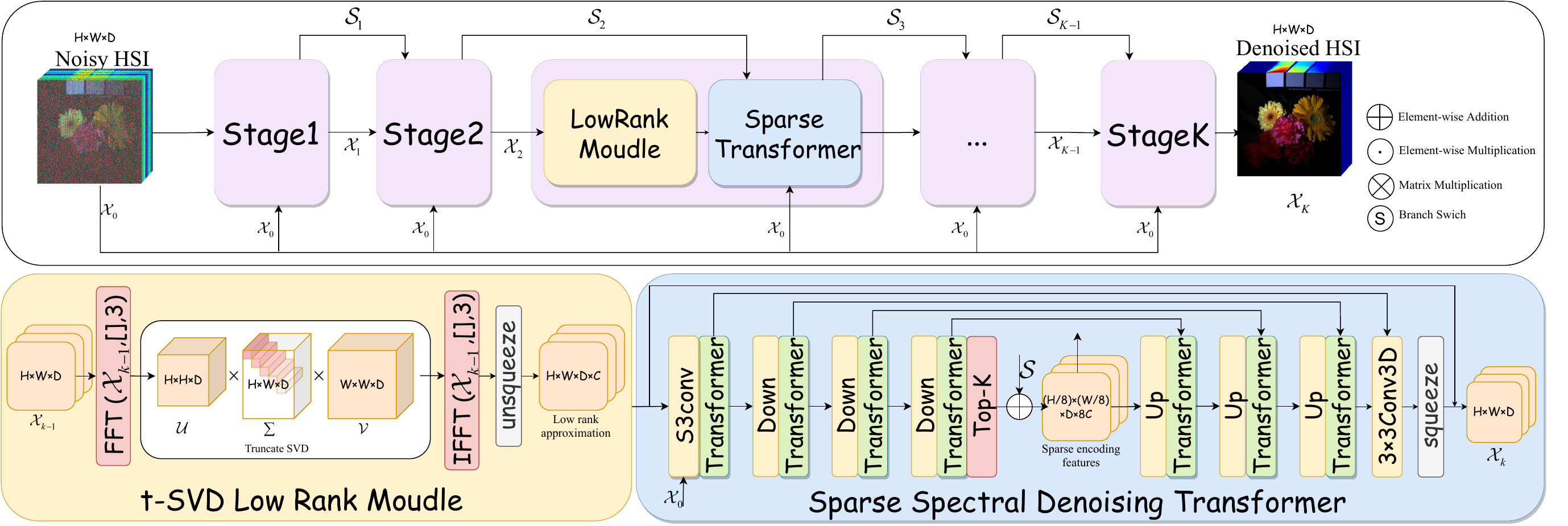}
    \caption{Overview of the DU-TRPCA network. (a) The complete architecture, consisting of $K$ stages, each alternating a thresholded t-SVD (low-rank) module and a Top-K sparse HSDT Transformer U-Net module. (b) Detailed structure of the thresholded t-SVD module for adaptive low-rank approximation. (c) Architecture of the Top-K sparsity module, which integrates the HSDT Transformer U-Net for efficient sparse feature modeling and nonlinear denoising.
  }
  \label{fig:architecture}
\end{figure*}

\section{Related Work}\label{sec:related_work}

\subsection{Model-Driven Approaches}
Model-driven HSI denoising approaches leverage analytical priors such as low-rankness, sparsity, TV, and non-local self-similarity. Early works by Candès et al. \cite{xu2017denoising} and Wright et al. \cite{peng2022hyperspectral} introduced robust principal component analysis (RPCA), which separates low-rank and sparse components in matrices. For hyperspectral images, Lu et al.\cite{luTensorRobustPrincipal2020} proposed TRPCA, extending RPCA to higher-order tensors by introducing tensor nuclear norms and corresponding optimization frameworks to better preserve the inherent multi-dimensional structure. Compared with conventional subspace and matrix/tensor decomposition methods, the TRPCA model based on t-SVD and the tensor nuclear norm can more effectively capture global spatial-spectral correlations and improve denoising performance under mixed noise conditions. In addition, models such as those by He et al. \cite{he2015total} and He et al. \cite{he2020non} further enhance denoising results by incorporating TV regularization and non-local self-similarity priors, respectively. However, these optimization-based models remain sensitive to prior assumptions and hyperparameter tuning, and their performance may degrade when confronted with complex real-world noise. 

\subsection{Data-Driven Approaches}
Data-driven methods based on deep learning address HSI denoising by directly learning mappings from noisy to clean data. Early CNNs, such as HSID-CNN~\cite{yuan2018hyperspectral}, mainly performed band-wise or band-group processing, with limited spectral interaction. Subsequent works, including 3D convolutions~\cite{sidorov2019deep} and quasi-recurrent pooling~\cite{wei3DQuasirecurrentNeural2021}, improved spatial-spectral feature modeling. Recently, transformer-based architectures such as SST and SERT~\cite{liSpatialspectralTransformerHyperspectral2023, liSpectralEnhancedRectangle2023} further enhanced denoising performance by jointly modeling spatial and spectral dependencies. HSDT integrates separable convolutions, guided spectral attention, and a self-modulated feed-forward network, flexibly accommodating arbitrary numbers of spectral bands and achieving state-of-the-art results, making it well suited for real-world HSI data with varying spectral configurations\cite{laiHybridSpectralDenoising2023}. 

Despite these advances, purely data-driven models often require extensive paired data, lack interpretability, and may generalize poorly to unseen conditions. These limitations have spurred interest in hybrid approaches that combine data-driven learning with model-based priors.

Recent work in low-level vision has explored integrating sparsity constraints, such as Top-K sparse attention, into transformer architectures, typically applying them to attention maps for improved feature selection and robustness to structured noise~\cite{chen2023learning,zhou2024adapt}. Building on these strengths, we integrate a sparsity-regularized HSDT module into a deep unfolding framework, explicitly combining model-driven priors with the strong representational capabilities of transformers for more robust and generalizable HSI denoising. Notably, unlike existing approaches where Top-K sparsity is applied to the attention maps, our method imposes Top-K sparsity directly on the feature representations. Notably, unlike previous approaches that apply Top-K sparsity to attention maps, our method imposes Top-K sparsity directly on feature representations. This provides a more explicit correspondence to the sparse regularization in TRPCA theory, ensuring greater interpretability and theoretical grounding.

\subsection{Hybrid Paradigms}
To leverage the complementary strengths of model-driven and data-driven approaches, hybrid HSI denoising methods integrate analytic priors into deep learning frameworks. A prominent direction is the PnP paradigm. For example, Deep-PnP~\cite{laiDeepPlugandplayPrior2022} replaces the proximal operator in ADMM with a fixed pretrained denoiser, thus retaining convergence guarantees. However, since the denoiser is not jointly optimized with the reconstruction objective and remains unchanged during inference, the model's adaptability to different or unknown noise patterns is inherently limited. Another line of research embeds prior-driven modules within end-to-end architectures. T3SC~\cite{bodrito2021trainable} introduces a learnable sparse coding block, LR-Net~\cite{zhangLRNetLowRankSpatialSpectral2021} imposes low-rank constraints on feature tensors, MAC-Net~\cite{xiongMACNetModelAidedNonlocal2022} integrates non-local filtering with deep priors, and HyLoRa~\cite{tanLowrankPromptguidedTransformer2024} employs low-rank prompts to guide transformers. Although these models structurally encode domain knowledge, the analytic priors typically function as soft constraints during end-to-end training, which can diminish interpretability and reduce the influence of the priors. Moreover, most existing architecture-guided designs either focus on a single prior or apply multiple priors sequentially in a fixed manner, rather than enforcing strict stage-wise alternation. This limits their ability to effectively address complex mixed noise scenarios in which both global structure and local outliers must be simultaneously modeled and removed.

Deep unfolding networks offer a principled framework for incorporating priors into learnable architectures by mapping each iteration of an optimization algorithm onto a trainable network layer~\cite{zhangISTAnetInterpretableOptimizationinspired2018,alhejailiRecursionsAreAll2023}. This approach enables explicit, interpretable stage-wise alternation between different modules, theoretically allowing for the strict coupling of complementary priors.  
Several deep unfolding models have recently been developed for HSI denoising, each with distinct limitations. ILR-Net~\cite{yeIterativeLowrankNetwork2024} unfolds an iterative low-rank optimization process, yet focuses exclusively on low-rank modeling and does not explicitly model sparse or local outlier structures. DNA-Net~\cite{zengDegradationnoiseawareDeepUnfolding2025} incorporates transformer modules into the unfolding scheme, but replaces explicit model-based priors with data-driven attention, reducing interpretability and weakening the physical grounding of the model. SMDS-Net~\cite{xiongSMDSnetModelGuided2022} combines a one-shot spectral projection with iterative sparse coding blocks, but lacks true stage-wise alternation between low-rank and sparse priors, limiting their mutual refinement and the capacity to address mixed noise scenarios.

In summary, existing deep unfolding approaches for HSI denoising tend to focus on either low-rank or data-driven modules in isolation, or couple them in an imbalanced or non-alternating manner. None has achieved a framework with explicit, rigorous stage-wise alternation between low-rank and sparse modules—an interaction that is critical for robust removal of complex mixed noise. This motivates the design of our proposed DU-TRPCA, which strictly alternates thresholded t-SVD low-rank and Top-K sparse transformer modules within a deep unfolding scheme, thereby achieving tighter integration of complementary priors, enhanced interpretability, and improved denoising under challenging conditions.

\section{Method}

\subsection{Problem Formulation and Alternating Optimization}
Following \cite{luTensorRobustPrincipal2020}, the classical TRPCA decomposes an observed tensor $\mathcal{X}$ into low-rank ($\mathcal{L}$) and sparse ($\mathcal{S}$) components, formulated as:

\begin{equation}
\min_{\mathcal{L}, \mathcal{S}} \|\mathcal{L}\|_{*} + \lambda\|\mathcal{S}\|_{1}, \quad \text{s.t.} \quad \mathcal{X} = \mathcal{L} + \mathcal{S}
\end{equation}

where $\|\mathcal{L}\|_{*}$ denotes the tensor nuclear norm based on t-SVD, and $\|\mathcal{S}\|_{1}$ indicates the element-wise $\ell_1$ norm.

To solve this convex optimization problem, TRPCA adopts an alternating minimization approach with iterative updates:

\begin{equation}
\left\{
\begin{aligned}
\mathcal{L}^{k+1} &= \arg\min_{\mathcal{L}} \frac{1}{2}\|\mathcal{X} - \mathcal{L} - \mathcal{S}^{k}\|_{F}^{2} + \lambda_{L}\|\mathcal{L}\|_{*}, \\
\mathcal{S}^{k+1} &= \arg\min_{\mathcal{S}} \frac{1}{2}\|\mathcal{X} - \mathcal{L}^{k+1} - \mathcal{S}\|_{F}^{2} + \lambda_{S}\|\mathcal{S}\|_{1}.
\end{aligned}
\right.
\end{equation}

\subsection{Deep Unfolding Network Architecture}
To integrate the interpretability and convergence stability of TRPCA with the adaptability of deep learning, we design DU-TRPCA, a deep unfolding network that explicitly alternates between low-rank and sparse updates at each unfolded iteration, directly inspired by the iterative optimization framework above. Specifically, DU-TRPCA consists of two carefully designed modules:

\paragraph{Low-Rank Module via Differentiable Thresholded t-SVD}

Motivated by the TRPCA framework~\cite{luTensorRobustPrincipal2020}, our low-rank module promotes global spatial-spectral correlations by enforcing a low tubal rank at each iteration. Given the current sparse estimate $\mathcal{S}^k$, the low-rank component $\mathcal{L}^{k+1}$ is updated by solving:

\begin{equation}
\mathcal{L}^{k+1} = \arg\min_{\mathcal{L}} \frac{1}{2}\|\mathcal{X} - \mathcal{L} - \mathcal{S}^{k}\|_{F}^{2} + \lambda_{L}\|\mathcal{L}\|_{*},
\end{equation}

where $\|\mathcal{L}\|_{*}$ denotes the tensor nuclear norm induced by t-SVD. The closed-form solution is obtained by a rank-$r$ truncated t-SVD projection:
\begin{equation}
\mathcal{L}^{k+1} = \mathcal{X} - r_L^k \cdot \left(\mathcal{X} - \mathcal{P}_{\text{t-SVD},r}(\mathcal{X} - \mathcal{S}^k)\right),
\end{equation}
 where $r_L^k$ is a learnable residual weight and $\mathcal{P}_{\text{t-SVD},r}(\cdot)$ denotes rank-$r$ truncated t-SVD projection:
\begin{equation}
\left\{
\begin{aligned}
&\bar{\mathcal{Y}} = \operatorname{fft}(\mathcal{X} - \mathcal{S}^{k}, [], 3) \\[0.5em]
&\bar{\mathcal{Y}}^{(i)}_r = \operatorname{Truncate}_r(\operatorname{SVD}(\bar{\mathcal{Y}}^{(i)})), \quad i=1,\ldots,n_3 \\[0.5em]
&\mathcal{L}^{k+1} = \operatorname{ifft}(\{\bar{\mathcal{Y}}^{(i)}_r\}_{i=1}^{n_3}, [], 3)
\end{aligned}
\right.
\end{equation}
Here, $\operatorname{Truncate}_r(\cdot)$ retains only the largest $r$ singular values and their corresponding vectors for each frontal slice in the frequency domain, effectively performing hard thresholding of the tubal singular values~\cite[Algorithm 2]{luTensorRobustPrincipal2020}.

However, hard truncation is inherently non-differentiable, which obstructs standard gradient propagation in deep networks. To address this, we implement a custom backward process: during the forward pass, we perform strict hard thresholded t-SVD as described above; during the backward pass, gradients are propagated only through the preserved singular components (i.e., the top-$r$ singular vectors and values), while the gradients with respect to the discarded components and the imaginary parts in the frequency domain are set to zero.  This pseudo-gradient approach provides a numerically stable and computationally efficient surrogate for the true gradient, thereby supporting effective end-to-end training of our deep unfolding network. This practical treatment is inspired by recent advances in differentiable SVD for deep learning~\cite{ionescu2015matrix, yeIterativeLowrankNetwork2024}, but is tailored to our specific tensor structure and engineering considerations.

This module explicitly enforces a low-rank prior at each unfolding stage, robustly capturing the global spatial-spectral structure of HSIs. By alternating this operation with the sparse module, our network achieves interpretable and effective separation of principal components and structured noise, consistent with the TRPCA framework.

\paragraph{Sparse Module via Top-K Sparse Transformer}

\begin{figure}[!t]           
  \centering
  \includegraphics[width=\columnwidth]{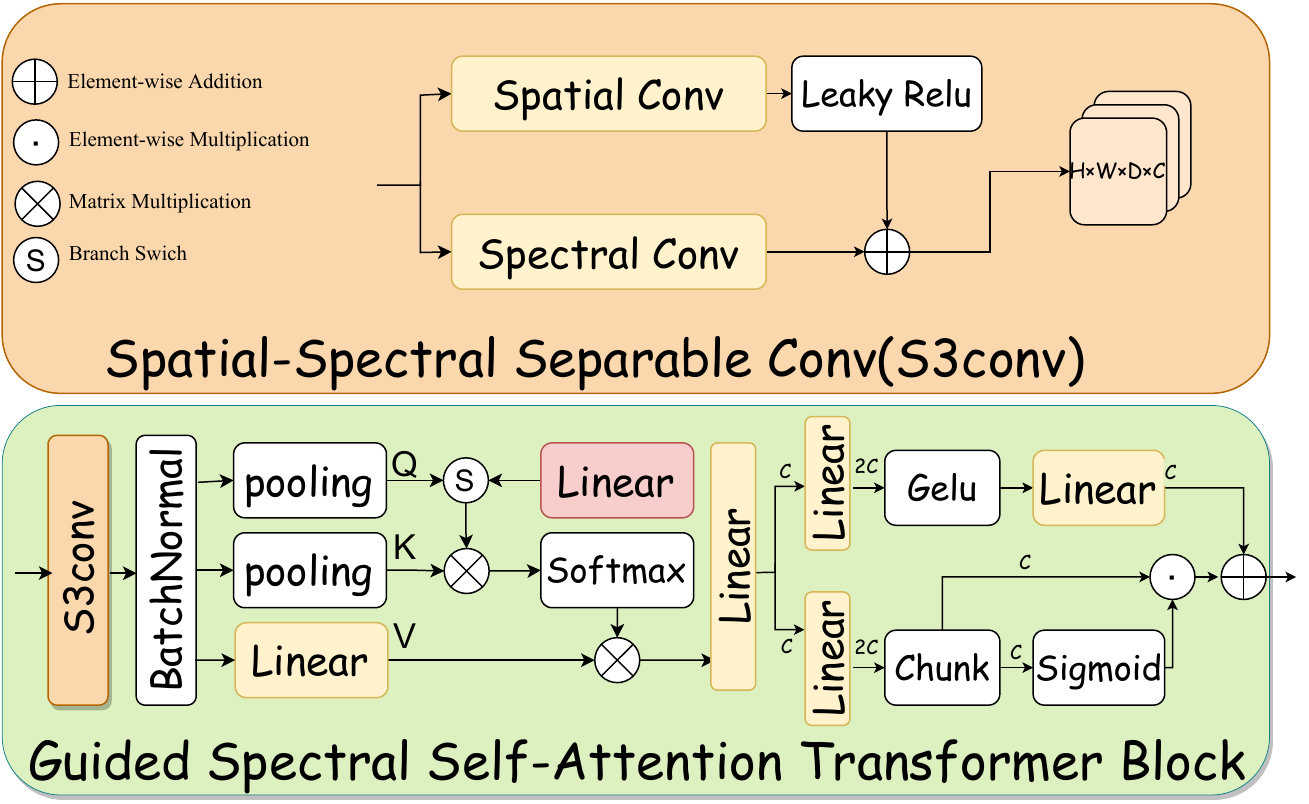}
  \caption{
Architecture of the Guided Spectral Self-Attention (GSSA) transformer block~\cite{laiHybridSpectralDenoising2023}. 
(a) Spatial-Spectral Separable Convolution (S3Conv). 
(b) Guided Spectral Self-Attention transformer block, comprising GSSA and self-modulated feed-forward network (SM-FFN).
}
  \label{fig:architecture}
\end{figure}

In classical TRPCA, the sparse component is updated at each iteration by solving the following subproblem:
\begin{equation}
\mathcal{S}^{k+1} = \arg\min_{\mathcal{S}} \frac{1}{2} \|\mathcal{X} - \mathcal{L}^{k+1} - \mathcal{S}\|_F^2 + \lambda_S \|\mathcal{S}\|_1,
\end{equation}
which admits a closed-form solution via element-wise soft-thresholding.

Inspired by this proximal mapping, we generalize the sparse update using a neural network module based on a transformer with adaptive Top-K sparsity:
\begin{equation}
\mathcal{S}^{k+1} = \mathcal{X} - r_S^k \cdot \left(\mathcal{X} - \mathcal{P}_{\text{Top-K}}(\mathcal{X} - \mathcal{L}^{k+1})\right),
\end{equation}
where $r_S^k$ is a learnable residual weight and $\mathcal{P}_{\text{Top-K}}(\cdot)$ integrates a standard transformer encoder with an additional Top-K sparsity constraint at its deepest encoder layer. This sparsity constraint explicitly mimics the TRPCA sparse regularization, selecting and retaining only the most salient spatial-spectral features corresponding to significant sparse noise elements. Unlike conventional $\ell_1$-norm soft-thresholding, our Top-K sparse transformer provides an adaptive, data-driven realization of the sparse prior, enabling effective removal of complex structured noise and local anomalies.

The detailed procedure is as follows:
\begin{equation}
\left\{
\begin{aligned}
&\mathcal{E} = \mathcal{T}_{\mathrm{enc}}\left(\mathcal{X} - \mathcal{L}^{k+1}\right) \\[0.5em]
&\mathcal{E}_{\mathrm{Top}-K} = \operatorname{TopK}\left(\mathcal{E}; k\right) \\[0.5em]
&\mathcal{S}^{k+1} = \mathcal{T}_{\mathrm{dec}}\left(\mathcal{E}_{\mathrm{Top}-K}\right)
\end{aligned}
\right.
\end{equation}

where $\mathcal{T}_{\mathrm{enc}}$ and $\mathcal{T}_{\mathrm{dec}}$ denote the transformer encoder and decoder (following the HSDT block design), and $\operatorname{TopK}(\cdot; k)$ imposes a learnable sparsity constraint at the feature bottleneck. The final output $\mathcal{S}^{k+1}$ resides in the same space as the observed data, thus strictly corresponding to the TRPCA sparse update. All modules and parameters, including the Top-K rate, are optimized end-to-end, enabling expressive and adaptive modeling of structured sparse noise.

\subsection{Loss Function}

Following DGUNet~\cite{mouDeepGeneralizedUnfolding2022}, we optimize our DU-TRPCA network using the standard $\ell_2$ loss, which involves the outputs from all unfolding stages. Specifically, for each training sample, given the degraded measurement $\mathcal{Y}$ and the ground-truth image $\mathcal{X}$, the loss is defined as:
\begin{equation}
    \mathcal{L}(\phi) = \sum_{k=1}^{K} \left\| \mathcal{X} - \hat{\mathcal{X}}^{k} \right\|_{F}^{2},
\end{equation}

where $K$ denotes the total number of unfolding stages, and $\hat{\mathcal{X}}^{k} $ represents the restoration result at the $k$-th stage. Here, $\phi$ denotes all trainable parameters of the network. This stage-wise supervision encourages all intermediate outputs to approximate the ground truth, facilitating stable convergence and improved denoising performance.

\sisetup{detect-mode,round-mode=places,round-precision=3,parse-numbers=false}

\begin{table*}[!htbp]
\caption{Experimental results under mixed noise types}
\label{tab:icvl-mixed}
\centering\scriptsize
\renewcommand{\arraystretch}{1.1}
\begin{adjustbox}{max width=\textwidth}
\begin{tabular}{
  l||
  S[table-format=2.3] S[table-format=1.3] S[table-format=1.3]|
  S[table-format=2.3] S[table-format=1.3] S[table-format=1.3]|
  S[table-format=2.3] S[table-format=1.3] S[table-format=1.3]|
  S[table-format=2.3] S[table-format=1.3] S[table-format=1.3]|
  S[table-format=2.3] S[table-format=1.3] S[table-format=1.3]
}
\toprule
\multirow{2}{*}{Method}
  & \multicolumn{3}{c|}{non‐iid}
  & \multicolumn{3}{c|}{stripe}
  & \multicolumn{3}{c|}{deadline}
  & \multicolumn{3}{c|}{impulse}
  & \multicolumn{3}{c}{mixture} \\
\cmidrule(lr){2-4}
\cmidrule(lr){5-7}
\cmidrule(lr){8-10}
\cmidrule(lr){11-13}
\cmidrule(lr){14-16}
  & {PSNR} & {SSIM} & {SAM}
  & {PSNR} & {SSIM} & {SAM}
  & {PSNR} & {SSIM} & {SAM}
  & {PSNR} & {SSIM} & {SAM}
  & {PSNR} & {SSIM} & {SAM} \\
\midrule
NGmeet (TPAMI’20)
  & 32.082 & 0.889 & 0.109
  & 32.044 & 0.889 & 0.110
  & 31.154 & 0.885 & 0.128
  & 25.376 & 0.755 & 0.473
  & 24.716 & 0.755 & 0.501 \\
LRTFDFR (TGRS’20)
  & 32.323 & 0.678 & 0.304
  & 31.285 & 0.639 & 0.334
  & 29.899 & 0.609 & 0.396
  & 30.044 & 0.664 & 0.337
  & 27.549 & 0.572 & 0.418 \\
HyMix (TNNLS’21)
  & 37.415 & 0.961 & 0.094
  & 33.631 & 0.888 & 0.151
  & 32.881 & 0.899 & 0.134
  & 29.183 & 0.847 & 0.487
  & 23.791 & 0.696 & 0.524 \\
GRUNet (Neurocomputing ’22)
  & 42.895 & 0.978 & 0.047
  & 42.397 & 0.977 & 0.049
  & 42.109 & 0.976 & 0.050
  & 40.703 & 0.966 & \textbf{0.067}
  & 38.510 & 0.957 & {\underline{0.081}} \\
FFDNet (Neurocomputing ’22)
  & 35.672 & 0.941 & 0.090
  & 35.542 & 0.938 & 0.091
  & 34.804 & 0.933 & 0.092
  & 26.013 & 0.752 & 0.216
  & 26.060 & 0.750 & 0.207 \\
QRNN3D (TNNLS’20)
  & 42.972 & 0.980 & 0.050
  & 42.670 & 0.979 & 0.051
  & 42.386 & 0.978 & 0.052
  & 40.409 & 0.957 & 0.094
  & 39.255 & 0.952 & 0.094 \\
T3SC (NIPS’21)
  & 41.287 & 0.973 & 0.066
  & 40.847 & 0.971 & 0.072
  & 39.543 & 0.966 & 0.096
  & 36.068 & 0.932 & 0.203
  & 34.477 & 0.922 & 0.228 \\
SMDSNet (TIP’22)
  & 33.955 & 0.941 & 0.123
  & 33.610 & 0.938 & 0.128
  & 32.910 & 0.934 & 0.141
  & 28.215 & 0.837 & 0.284
  & 26.891 & 0.825 & 0.315 \\
TRQ3D (RS’22)
  & 42.959 & 0.982 & 0.044
  & 42.664 & 0.981 & 0.046
  & 42.579 & 0.981 & 0.046
  & 40.881 & {\underline{0.968}} & {\underline{0.078}}
  & 39.876 & 0.964 & \textbf{0.078} \\
SST (AAAI’22)
  & 42.801 & 0.981 & 0.050
  & 42.489 & 0.980 & 0.051
  & 42.275 & 0.979 & 0.053
  & 39.936 & 0.960 & 0.082
  & 38.618 & 0.952 & 0.086 \\
SERT (CVPR’23)
  & 43.680 & 0.982 & 0.046
  & 43.405 & 0.981 & 0.048
  & 43.261 & 0.981 & 0.048
  & 40.287 & 0.960 & 0.088
  & 39.090 & 0.958 & 0.091 \\
HyLora (TGRS’24)
  & 38.603 & 0.954 & 0.068
  & 38.264 & 0.951 & 0.070
  & 37.783 & 0.950 & 0.072
  & 34.874 & 0.912 & 0.119
  & 34.099 & 0.910 & 0.113 \\
ILRNet (TGRS’24)
  & 43.180 & 0.979 & 0.049
  & 42.865 & 0.979 & 0.051
  & 42.827 & 0.978 & 0.050
  & 40.671 & 0.960 & 0.100
  & 39.600 & 0.956 & 0.106 \\
HSDT (ICCV’23)
  & 44.691 & 0.984 & 0.038
  & 44.429 & 0.984 & 0.039
  & 44.309 & 0.983 & 0.039
  & 41.050 & 0.959 & 0.110
  & 40.205 & 0.957 & 0.115 \\
HSDT\_L (ICCV’23)
  & \textbf{44.975} & {\underline{0.985}} & {\underline{0.036}}
  & {\underline{44.740}} & {\underline{0.984}} & {\underline{0.038}}
  & {\underline{44.632}} & {\underline{0.984}} & {\underline{0.037}}
  & {\underline{41.828}} & 0.965       & 0.102
  & {\underline{40.912}} & {\underline{0.962}} & 0.106 \\
\midrule
\textbf{Ours}
  & {\underline{44.883}} & \textbf{0.985} & \textbf{0.035}
  & \textbf{44.767} & \textbf{0.985} & \textbf{0.035}
  & \textbf{44.728} & \textbf{0.984} & \textbf{0.035}
  & \textbf{42.362} & \textbf{0.969} & 0.088
  & \textbf{41.960} & \textbf{0.968} & 0.086 \\
\bottomrule
\end{tabular}
\end{adjustbox}
\end{table*}

\graphicspath{{figures/denoise/}}
\begin{figure*}[ht]
  \centering
  \foreach \i/\imgfile/\algoname in {
    1/gt/Clean,
    2/noised/Noisy,
    3/NGmeet/NGmeet,
    4/LRTFDFR/LRTFDFR,
    5/Hymix/HyMix,
    6/GRUNet/GRUNet,
    7/FFDNet/FFDNet,
    8/qrnn3d/QRNN3D,
    9/t3sc/T3SC,
    10/smdsnet/SMDSNet,
    11/trq3d/TRQ3D,
    12/sst/SST,
    13/sert/SERT,
    14/hylora/HyLora,
    15/ilrnet/ILRNet,
    16/hsdt/HSDT,
    17/hsdt_l/HSDT\_L,
    18/ours/Ours
  }{
    \begin{subfigure}[t]{0.10\textwidth}
      \centering
      \begin{tikzpicture}[inner sep=0,
          spy using outlines={
            rectangle,
            magnification=4,
            size=0.55\linewidth,
            connect spies={draw=red,thick},
            every spy on node/.style={draw=red},
            every spy in node/.style={draw=red}
          }]
        \node (img) {\includegraphics[width=\linewidth]{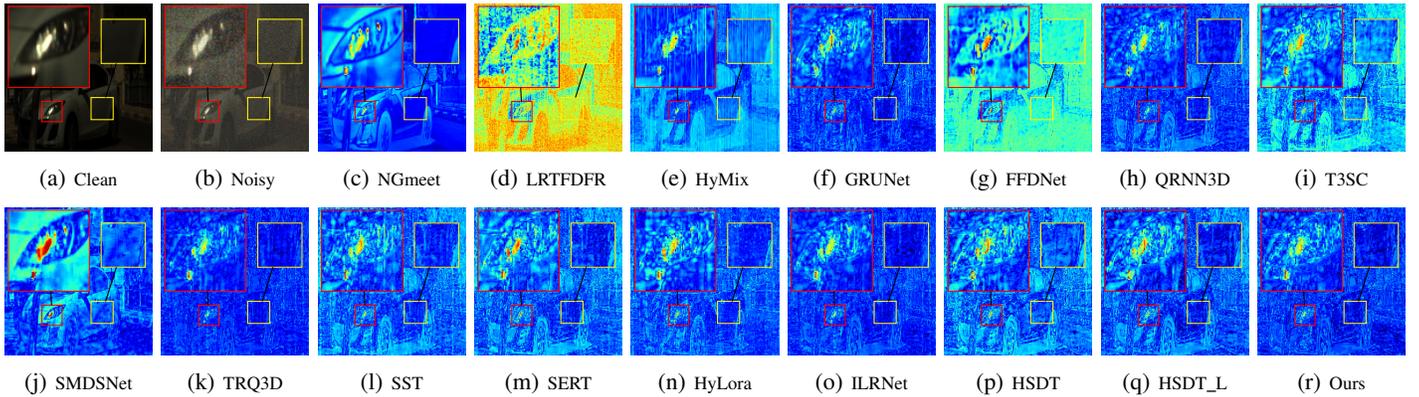}};
        
        \coordinate (region) at ($(img.center)+(-0.18\linewidth,-0.23\linewidth)$);
        \spy on (region)
            in node [anchor=north east,
                     xshift=-0.43\linewidth,
                     yshift=-0.02\linewidth]
            at (img.north east);

        \coordinate (region2) at ($(img.center)+(0.16\linewidth,-0.21\linewidth)$);
        \spy[
          magnification=2,
          size=0.3\linewidth,
          connect spies={draw=yellow, thick},
          every spy on node/.style={draw=yellow},
          every spy in node/.style={draw=yellow}
        ]
            on (region2)
        in node [anchor=south east,
                 xshift=-0.05\linewidth,  
                 yshift= 0.6\linewidth]  
            at (img.south east);

      \end{tikzpicture}
      \caption{\scriptsize \algoname}
      \label{sub:\imgfile}
    \end{subfigure}%
    \ifnum\i=9 \par\medskip\fi
  }
  \caption{Denoising comparison on the mixture‐noise scenario of ICVL’s \texttt{Labtest\_0910-1510}.}
  \label{fig:icvl_mixture_2rows}
\end{figure*}

\graphicspath{{figures/reflectance/}}
\begin{figure*}[!htbp]
  \captionsetup[subfigure]{%
    font=scriptsize,        
    labelfont=bf,           
    skip=1pt,               
    justification=centering 
  }
  \centering
  \foreach \i/\imgfile/\algoname in {
    1/gt/Clean,
    2/noised/Noisy,
    3/NGmeet/NGmeet,
    4/LRTFDFR/LRTFDFR,
    5/Hymix/HyMix,
    6/GRUNet/GRUNet,
    7/FFDNet/FFDNet,
    8/qrnn3d/QRNN3D,
    9/t3sc/T3SC,
    10/smdsnet/SMDSNet,
    11/trq3d/TRQ3D,
    12/sst/SST,
    13/sert/SERT,
    14/hylora/HyLora,
    15/ilrnet/ILRNet,
    16/hsdt/HSDT,
    17/hsdt_l/HSDT\_L,
    18/ours/Ours
  }{
    \begin{subfigure}[t]{0.105\textwidth}  
      \centering
      \includegraphics[width=\linewidth]{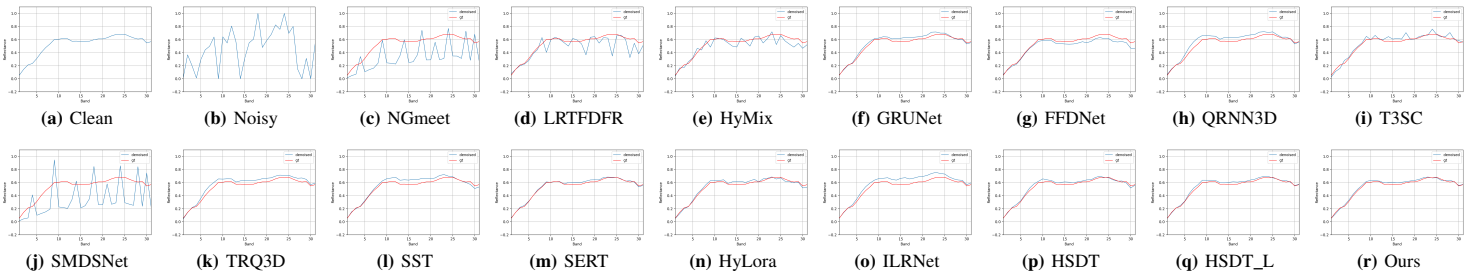}
      \caption{\algoname}
      \label{sub:\imgfile}
    \end{subfigure}%
    \ifnum\i=9 \par\medskip\fi
  }
  \caption{
Comparison of reflectance spectra at pixel $(161,376)$ in the ICVL dataset’s Labtest\_0910-1510 HSI. Results of all denoising methods are shown, with ground truth in \textcolor{red}{red}.
}

  \label{fig:icvl_mixture_2rows}
\end{figure*}

\subsection{Stage-wise Alternation and End-to-End Training}
DU-TRPCA alternates the low-rank and sparse modules in a stage-wise manner, closely mirroring the iterative updates of classical TRPCA. Each iteration is unfolded into a network stage, preserving the optimization structure:
\begin{equation}
\left\{
\begin{aligned}
\mathcal{L}^{k+1} &= \mathcal{X} - r_L^k \cdot \left(\mathcal{X} - \mathcal{P}_{\text{t-SVD},r}(\mathcal{X} - \mathcal{S}^k)\right), \\
\mathcal{S}^{k+1} &= \mathcal{X} - r_S^k \cdot \left(\mathcal{X} - \mathcal{P}_{\text{Top-K}}(\mathcal{X} - \mathcal{L}^{k+1})\right)
\end{aligned}
\right.
\end{equation}
All module parameters are jointly optimized end-to-end, using supervised denoising objectives. This design tightly couples the two complementary priors and enables expressive, adaptive modeling within an interpretable optimization-inspired framework. By embedding the proximal operators of TRPCA into a deep unfolding network, DU-TRPCA inherits the theoretical guarantees, convergence behavior, and interpretability of the original optimization framework. The low-rank module retains the robustness and global structure modeling of t-SVD-based TRPCA, while the learnable Top-K sparse module extends classical soft-thresholding to an adaptive, data-driven regime. This synergy yields a principled, interpretable, and effective solution for challenging mixed-noise HSI denoising tasks.

\section{Experiments}

We perform comprehensive experiments on synthetic and real-world hyperspectral datasets to validate the effectiveness of our approach. The experimental setup includes comparisons with state-of-the-art methods under various noise conditions, as well as ablation studies to investigate the impact of different components of our model.

\begin{table*}[!htbp]
\caption{Experimental results under mixed noise types on CAVE dataset}
\label{tab:cave-mixed}
\centering\scriptsize
\renewcommand{\arraystretch}{1.1}
\begin{adjustbox}{max width=\textwidth}
\begin{tabular}{
  l||
  S[table-format=2.3] S[table-format=1.3] S[table-format=1.3]|
  S[table-format=2.3] S[table-format=1.3] S[table-format=1.3]|
  S[table-format=2.3] S[table-format=1.3] S[table-format=1.3]|
  S[table-format=2.3] S[table-format=1.3] S[table-format=1.3]|
  S[table-format=2.3] S[table-format=1.3] S[table-format=1.3]
}
\toprule
\multirow{2}{*}{Method}
  & \multicolumn{3}{c|}{non‐iid}
  & \multicolumn{3}{c|}{stripe}
  & \multicolumn{3}{c|}{deadline}
  & \multicolumn{3}{c|}{impulse}
  & \multicolumn{3}{c}{mixture} \\
\cmidrule(lr){2-4}
\cmidrule(lr){5-7}
\cmidrule(lr){8-10}
\cmidrule(lr){11-13}
\cmidrule(lr){14-16}
  & {PSNR} & {SSIM} & {SAM}
  & {PSNR} & {SSIM} & {SAM}
  & {PSNR} & {SSIM} & {SAM}
  & {PSNR} & {SSIM} & {SAM}
  & {PSNR} & {SSIM} & {SAM} \\
\midrule
NGmeet (TPAMI’20)
  & 28.940 & 0.804 & 0.398
  & 29.028 & 0.801 & 0.431
  & 28.301 & 0.800 & 0.407
  & 22.567 & 0.603 & 0.765
  & 21.805 & 0.581 & 0.792 \\

LRTFDFR (TGRS’20)
  & 25.506 & 0.377 & 0.819
  & 24.360 & 0.328 & 0.861
  & 25.212 & 0.372 & 0.849
  & 25.408 & 0.392 & 0.745
  & 25.289 & 0.395 & 0.761 \\

Hymix (TNNLS’21)
  & 36.488 & 0.902 & 0.292
  & 34.145 & 0.834 & 0.380
  & 34.114 & 0.877 & 0.308
  & 28.918 & 0.739 & 0.743
  & 25.354 & 0.622 & 0.775 \\

GRUNet (Neurocomputing ’22)
  & 37.793 & 0.937 & 0.279
  & 37.083 & 0.928 & 0.294
  & 37.279 & 0.935 & 0.283
  & 35.494 & 0.888 & 0.398
  & 34.599 & 0.887 & 0.390 \\

FFDNet (Neurocomputing ’22)
  & 26.855 & 0.772 & 0.563
  & 26.851 & 0.771 & 0.563
  & 26.618 & 0.769 & 0.565
  & 22.177 & 0.505 & 0.640
  & 21.559 & 0.491 & 0.653 \\

QRNN3D (TNNLS’20)
  & 36.482 & 0.896 & 0.279
  & 35.958 & 0.885 & 0.295
  & 35.945 & 0.895 & 0.284
  & 33.150 & 0.775 & 0.380
  & 32.650 & 0.770 & 0.391 \\

T3SC (NIPS’21)
  & 38.154 & 0.926 & 0.232
  & 37.665 & 0.917 & 0.250
  & 37.283 & 0.918 & 0.259
  & 33.011 & 0.814 & 0.470
  & 37.283 & 0.918 & 0.259 \\

SMDSNet (TIP’22)
  & 31.188 & 0.854 & 0.462
  & 30.647 & 0.849 & 0.464
  & 30.946 & 0.860 & 0.461
  & 26.319 & 0.682 & 0.665
  & 25.168 & 0.665 & 0.688 \\

TRQ3D (RS’22)
  & 38.256 & 0.940 & 0.199
  & 37.739 & 0.931 & 0.214
  & 37.810 & 0.937 & 0.206
  & 35.394 & 0.863 & 0.308
  & 34.996 & 0.864 & 0.316 \\

SST (AAAI’22)
  & 35.140 & 0.888 & 0.274
  & 34.849 & 0.880 & 0.285
  & 34.689 & 0.888 & 0.279
  & 31.460 & 0.754 & 0.413
  & 30.773 & 0.745 & 0.399 \\

SERT (CVPR’23)
  & 36.393 & 0.911 & 0.238
  & 36.087 & 0.906 & 0.252
  & 35.751 & 0.911 & 0.245
  & 32.135 & 0.769 & 0.382
  & 31.528 & 0.763 & 0.378 \\

Hylora (TGRS’24)
  & 37.299 & 0.932 & 0.206
  & 36.958 & 0.926 & 0.218
  & 36.729 & 0.932 & 0.212
  & 33.552 & 0.813 & 0.330
  & 32.763 & 0.803 & 0.348 \\

ILRNet (TGRS’24)
  & 40.287 & {\underline{0.962}} & {\underline{0.142}}
  & 39.545 & {\underline{0.955}} & {\underline{0.157}}
  & 39.905 & {\underline{0.960}} & {\underline{0.144}}
  & {\underline{37.880}} & {\underline{0.916}} & \textbf{0.231}
  & {\underline{37.393}} & {\underline{0.921}} & {\underline{0.226}} \\

HSDT (ICCV’23)
  & {\underline{40.336}} & 0.958 & 0.171
  & {\underline{39.781}} & 0.953 & 0.178
  & {\underline{40.036}} & 0.958 & 0.170
  & 37.113 & 0.885 & 0.292
  & 37.143 & 0.898 & 0.289 \\

\midrule
\textbf{Ours}
  & \textbf{41.235} & \textbf{0.968} & \textbf{0.116}
  & \textbf{40.662} & \textbf{0.964} & \textbf{0.127}
  & \textbf{41.005} & \textbf{0.967} & \textbf{0.115}
  & \textbf{38.674} & \textbf{0.928} & {\underline{0.237}}
  & \textbf{38.759} & \textbf{0.940} & \textbf{0.193} \\

\bottomrule
\end{tabular}
\end{adjustbox}
\end{table*}

\subsection{Experimental Settings}

\textit{1) Compared Methods:}
For comprehensive and fair evaluation, we compare our method with a wide range of state-of-the-art open-source hyperspectral image denoising algorithms, covering model-driven, data-driven, and hybrid-driven paradigms. Specifically, NGMeet~\cite{he2020non}, LRTFDFR~\cite{zheng2020double}, and HyMix~\cite{zhuang2021fasthymix} are selected as model-driven methods. QRNN3D~\cite{wei3DQuasirecurrentNeural2021}, TRQ3D~\cite{pang2022trq3dnet}, HSDT~\cite{laiHybridSpectralDenoising2023}, SST~\cite{liSpatialspectralTransformerHyperspectral2023}, and SERT~\cite{liSpectralEnhancedRectangle2023} represent data-driven approaches. Hybrid-driven methods are further divided into three categories: plug-and-play frameworks (FFDNet, GRUNet~\cite{laiDeepPlugandplayPrior2022}), architecture-guided designs (T3SC~\cite{bodrito2021trainable}, HyLoRa~\cite{tanLowrankPromptguidedTransformer2024}), and deep unfolding networks (SMDSNet~\cite{xiongSMDSnetModelGuided2022}, ILRNet~\cite{yeIterativeLowrankNetwork2024}).

\textit{2) Datasets:}
Experiments are conducted on both synthetic and real-world HSI datasets. For synthetic experiments, we strictly follow the widely adopted QRNN3D protocol~\cite{wei3DQuasirecurrentNeural2021}, utilizing the ICVL~\cite{arad2016sparse} and CAVE~\cite{yasuma2010generalized} benchmarks. The ICVL dataset, acquired by the Specim PS Kappa DX4 hyperspectral camera, consists of 201 indoor and outdoor scenes at a spatial resolution of $1392 \times 1300$ with 31 spectral bands (400--700~nm, 10~nm intervals). According to standard practice, 100 images are used for training (cropped into 50,000 patches of $64 \times 64 \times 31$ via overlap and augmentation), and 51 images for testing ($512 \times 512 \times 31$). The CAVE dataset, collected at Columbia University, comprises 32 indoor scenes with 31 bands. We select 8 images for training (2,400 patches of $64 \times 64 \times 31$) and 12 images for testing ($512 \times 512 \times 31$), consistent with QRNN3D. 

For real-noise evaluation, two widely-used remote sensing datasets are employed: Urban and Indian Pines. The Urban dataset ($307 \times 307 \times 210$), captured by the HYDICE sensor, depicts a real urban scene with spatially-variant and spectrally-correlated noise (e.g., stripes, dead pixels, deadlines, and impulse noise). The Indian Pines dataset ($144 \times 144 \times 220$), collected by the AVIRIS sensor, includes agricultural and forest regions with typical real-world noise and atmospheric effects.

To ensure experimental consistency, all data-driven and hybrid-driven methods are trained and evaluated on the same dataset splits, strictly following the QRNN3D protocol. Open-source weights are adopted when available; otherwise, models are retrained from scratch with identical settings, using the parameter configurations reported in the respective original papers, to guarantee fairness.

\textit{3) Noise Patterns:}
To comprehensively assess model robustness, we consider a diverse set of challenging noise scenarios on synthetic datasets: (\emph{i}) \textbf{Mixed Noise}, encompassing non-i.i.d. Gaussian, vertical stripes (stripe noise), deadlines (dead pixel lines), impulse, and their mixtures; and (\emph{ii}) \textbf{Gaussian Noise}, with additive white Gaussian noise of standard deviations $\sigma = 30, 50, 70$, as well as a blind setting where $\sigma$ is randomly sampled from $[30, 70]$ for each test image.

\textit{4) Evaluation Metrics:}
For quantitative assessment, we report peak signal-to-noise ratio (PSNR), structural similarity index (SSIM), and spectral angle mapper (SAM), which are standard in the HSI denoising literature. Higher PSNR and SSIM values, and lower SAM values, indicate better denoising performance.

\textit{5) Implementation Details:}
All deep learning models are implemented in PyTorch and trained on a workstation equipped with an Intel Xeon Platinum 8480+ CPU and Nvidia H800 GPU, using a batch size of 4. The Adam optimizer is employed with an initial learning rate of $1 \times 10^{-3}$ and multi-step scheduling, training for 80 epochs. For fair comparison, both the inference of deep learning models and all traditional methods are conducted on a desktop with Intel i5-12600KF CPU and Nvidia 4060Ti GPU (using PyTorch and MATLAB, respectively).

Following common practice in deep unfolding networks~\cite{mouDeepGeneralizedUnfolding2022}, our DU-TRPCA is implemented with a stage-wise parameter sharing scheme to balance model capacity and efficiency. Specifically, the parameters of the first stage are independently learned, while all subsequent stages share a common set of parameters. This “1+N” configuration allows the first stage to flexibly adapt to the input and provides sufficient representational power, whereas the parameter sharing in later stages substantially reduces the overall parameter count and memory footprint without compromising denoising performance.

\begin{table*}[!htbp]
\caption{Experimental results under Gaussian noise on ICVL dataset}
\label{tab:gauss_icvl}
\centering\scriptsize
\renewcommand{\arraystretch}{1.1}
\begin{adjustbox}{max width=\textwidth}
\begin{tabular}{
  l||
  S[table-format=2.2] S[table-format=3.2]||
  S[table-format=2.3] S[table-format=1.3] S[table-format=1.3]|
  S[table-format=2.3] S[table-format=1.3] S[table-format=1.3]|
  S[table-format=2.3] S[table-format=1.3] S[table-format=1.3]|
  S[table-format=2.3] S[table-format=1.3] S[table-format=1.3]
}
\toprule
\multirow{2}{*}{Method}
  & \multicolumn{1}{c|}{Params (M)}
  & \multicolumn{1}{c||}{Time (s)}
  & \multicolumn{3}{c|}{$\sigma=30$}
  & \multicolumn{3}{c|}{$\sigma=50$}
  & \multicolumn{3}{c|}{$\sigma=70$}
  & \multicolumn{3}{c}{blind} \\
\cmidrule(lr){2-2}
\cmidrule(lr){3-3}
\cmidrule(lr){4-6}
\cmidrule(lr){7-9}
\cmidrule(lr){10-12}
\cmidrule(lr){13-15}
  & {Params} & {Time}
  & {PSNR} & {SSIM} & {SAM}
  & {PSNR} & {SSIM} & {SAM}
  & {PSNR} & {SSIM} & {SAM}
  & {PSNR} & {SSIM} & {SAM} \\
\midrule
NGmeet (TPAMI'20)
  & --    & 182.68
  & 35.263 & 0.911 & 0.063
  & 34.619 & 0.901 & 0.068
  & 34.166 & 0.893 & 0.074
  & 35.020 & 0.906 & 0.065 \\
LRTFDFR (TGRS'20)
  & --    &  40.83
  & 35.116 & 0.780 & 0.214
  & 28.649 & 0.532 & 0.375
  & 27.128 & 0.652 & 0.248
  & 29.899 & 0.609 & 0.396 \\
Hymix (TNNLS'21)
  & --    &   2.05
  & 37.876 & 0.963 & 0.084
  & 35.045 & 0.933 & 0.119
  & 33.070 & 0.903 & 0.152
  & 36.993 & 0.949 & 0.100 \\
GRUNet (Neurocomputing '22)
  & 14.20 &   1.31
  & 43.107 & 0.974 & 0.051
  & 40.656 & 0.960 & 0.066
  & 37.212 & 0.911 & 0.118
  & 40.354 & 0.947 & 0.067 \\
FFDNet (Neurocomputing '22)
  &  0.49 &   0.01
  & 37.597 & 0.953 & 0.095
  & 36.340 & 0.932 & 0.103
  & 34.945 & 0.901 & 0.119
  & 36.634 & 0.939 & 0.103 \\
QRNN3D (TNNLS'20)
  &  0.86 &   0.58
  & 42.220 & 0.973 & 0.062
  & 40.145 & 0.959 & 0.074
  & 38.302 & 0.941 & 0.094
  & 41.371 & 0.966 & 0.069 \\
T3SC (NIPS'21)
  &  0.83 &   0.90
  & 42.459 & 0.971 & 0.074
  & 40.563 & 0.958 & 0.082
  & 39.161 & 0.947 & 0.090
  & 41.626 & 0.965 & 0.079 \\
SMDSNet (TIP'22)
  &  0.01 &   4.60
  & 41.721 & 0.969 & 0.063
  & 34.950 & 0.938 & 0.128
  & 33.949 & 0.922 & 0.133
  & 38.239 & 0.955 & 0.095 \\
TRQ3D (RS'22)
  &  0.68 &   0.86
  & 41.927 & 0.974 & 0.059
  & 40.141 & 0.962 & 0.066
  & 38.756 & 0.950 & 0.074
  & 41.175 & 0.969 & 0.062 \\
SST (AAAI'22)
  &  4.14 &   2.56
  & 43.783 & 0.978 & 0.046
  & 41.573 & 0.967 & 0.055
  & 40.047 & 0.956 & 0.063
  & 42.988 & 0.973 & 0.049 \\
SERT (CVPR'23)
  &  1.91 &   1.04
  & 44.012 & 0.979 & 0.045
  & 41.813 & 0.968 & 0.054
  & 40.249 & 0.957 & 0.065
  & 43.200 & 0.974 & 0.049 \\
HyLora (TGRS'24)
  &  3.15 &   1.35
  & 44.270 & \textbf{0.980} & 0.041
  & 42.054 & {\underline{0.969}} & 0.050
  & 40.513 & {\underline{0.959}} & 0.058
  & 43.478 & \textbf{0.975} & 0.045 \\
ILRNet (TGRS'24)
  &  3.67 &   2.05
  & 43.897 & 0.978 & 0.042
  & 41.767 & 0.967 & 0.048
  & 40.291 & 0.957 & 0.054
  & 43.173 & 0.973 & 0.045 \\
HSDT (ICCV'23)
  &  0.52 &   1.26
  & 44.025 & 0.978 & 0.042
  & 41.826 & 0.968 & 0.049
  & 40.345 & 0.958 & 0.055
  & 43.326 & 0.974 & 0.045 \\
HSDT\_L (ICCV'23)
  &  2.09 &   1.35
  & \textbf{44.308} & {\underline{0.979}} & 0.041
  & {\underline{42.101}} & {\underline{0.969}} & 0.048
  & {\underline{40.601}} & {\underline{0.959}} & 0.054
  & {\underline{43.597}} & 0.974 & 0.043 \\
\midrule
ours(-TopK)
  &  1.05 &   6.19
  & {\underline{44.302}} & 0.969 & \textbf{0.039}
  & \textbf{42.106} & \textbf{0.969} & \textbf{0.044}
  & \textbf{40.636} & \textbf{0.960} & \textbf{0.049}
  & \textbf{43.612} & {\underline{0.975}} & \textbf{0.041} \\
ours
  &  1.05 &   6.30
  & 44.081 & 0.978 & {\underline{0.040}}
  & 41.826 & 0.967 & {\underline{0.047}}
  & 40.339 & 0.958 & {\underline{0.052}}
  & 43.365 & 0.973 & {\underline{0.042}} \\
\bottomrule
\end{tabular}
\end{adjustbox}
\end{table*}

\begin{table*}[!htbp]
\caption{Experimental results under Gaussian noise on CAVE dataset }
\label{tab:gauss_cave}
\centering\scriptsize
\renewcommand{\arraystretch}{1.1}
\begin{adjustbox}{max width=\textwidth}
\begin{tabular}{
  l||
  S[table-format=2.3] S[table-format=1.3] S[table-format=1.3]|
  S[table-format=2.3] S[table-format=1.3] S[table-format=1.3]|
  S[table-format=2.3] S[table-format=1.3] S[table-format=1.3]|
  S[table-format=2.3] S[table-format=1.3] S[table-format=1.3]
}
\toprule
\multirow{2}{*}{Method}
  & \multicolumn{3}{c|}{$\sigma=30$}
  & \multicolumn{3}{c|}{$\sigma=50$}
  & \multicolumn{3}{c|}{$\sigma=70$}
  & \multicolumn{3}{c}{blind} \\
\cmidrule(lr){2-4}
\cmidrule(lr){5-7}
\cmidrule(lr){8-10}
\cmidrule(lr){11-13}
  & {PSNR} & {SSIM} & {SAM}
  & {PSNR} & {SSIM} & {SAM}
  & {PSNR} & {SSIM} & {SAM}
  & {PSNR} & {SSIM} & {SAM} \\
\midrule
NGmeet (TPAMI’20)
  & 31.439 & 0.858 & 0.241
  & 31.037 & 0.843 & 0.260
  & 30.614 & 0.831 & 0.274
  & 31.403 & 0.857 & 0.247 \\

LRTFDFR (TGRS’20)
  & 30.565 & 0.593 & 0.572
  & 22.057 & 0.210 & 0.873
  & 18.623 & 0.239 & 0.813
  & 29.806 & 0.547 & 0.599 \\

Hymix (TNNLS’21)
  & 37.997 & 0.938 & 0.211
  & 35.474 & 0.894 & 0.279
  & 33.615 & 0.848 & 0.342
  & 37.904 & 0.932 & 0.216 \\

GRUNet (Neurocomputing ’22)
  & 37.067 & 0.924 & 0.290
  & 36.432 & 0.918 & 0.281
  & 34.129 & 0.863 & 0.393
  & 36.511 & 0.913 & 0.297 \\

FFDNet (Neurocomputing ’22)
  & 27.061 & 0.782 & 0.562
  & 26.837 & 0.766 & 0.561
  & 26.535 & 0.750 & 0.562
  & 26.848 & 0.775 & 0.564 \\

QRNN3D (TNNLS’20)
  & 38.537 & 0.942 & 0.203
  & 36.246 & 0.906 & 0.261
  & 33.698 & 0.830 & 0.347
  & 38.178 & 0.932 & 0.216 \\

T3SC (NIPS’21)
  & 39.688 & 0.953 & 0.164
  & 37.854 & 0.932 & 0.197
  & 36.430 & 0.910 & 0.229
  & 39.422 & 0.951 & 0.169 \\

SMDSNet (TIP’22)
  & 37.540 & 0.932 & 0.231
  & 31.678 & 0.865 & 0.454
  & 27.273 & 0.767 & 0.496
  & 35.530 & 0.912 & 0.283 \\

TRQ3D (RS’22)
  & 39.363 & 0.954 & 0.163
  & 37.413 & 0.933 & 0.199
  & 35.799 & 0.906 & 0.241
  & 39.133 & 0.951 & 0.166 \\

SST (AAAI’22)
  & 38.999 & 0.946 & 0.187
  & 36.906 & 0.919 & 0.221
  & 35.295 & 0.891 & 0.257
  & 38.740 & 0.942 & 0.192 \\

SERT (CVPR’23)
  & 39.787 & 0.954 & 0.167
  & 37.710 & 0.930 & 0.205
  & 36.088 & 0.904 & 0.242
  & 39.506 & 0.951 & 0.170 \\

HyLora (TGRS’24)
  & 39.931 & 0.955 & 0.167
  & 37.935 & 0.934 & 0.201
  & 36.354 & 0.910 & 0.236
  & 39.680 & 0.952 & 0.172 \\

ILRNet (TGRS’24)
  & 40.604 & 0.964 & 0.120
  & 38.512 & 0.948 & 0.139
  & 37.073 & 0.934 & 0.156
  & 40.516 & 0.963 & 0.120 \\

HSDT (ICCV’23)
  & 40.724 & {\underline{0.966}} & 0.120
  & 38.669 & 0.950 & 0.142
  & 37.209 & 0.935 & 0.164
  & 40.588 & {\underline{0.965}} & 0.121 \\

\midrule
ours(-TopK)
  & \textbf{41.034} & \textbf{0.967} & \textbf{0.111}
  & \textbf{39.025} & \textbf{0.953} & \textbf{0.126}
  & \textbf{37.630} & \textbf{0.941} & \textbf{0.140}
  & \textbf{40.949} & \textbf{0.966} & \textbf{0.111} \\

ours
  & {\underline{40.880}} & 0.966            & {\underline{0.114}}
  & {\underline{38.847}} & {\underline{0.951}} & {\underline{0.131}}
  & {\underline{37.421}} & {\underline{0.938}} & {\underline{0.146}}
  & {\underline{40.767}} & 0.965            & {\underline{0.114}} \\

\bottomrule
\end{tabular}
\end{adjustbox}
\end{table*}

\subsection{Results on Synthetic Noise}
\textit{1) Mixed Noise:}
Tables~\ref{tab:icvl-mixed} and~\ref{tab:cave-mixed} present quantitative denoising results of all competing methods under the most challenging mixture noise scenarios. Data-driven approaches consistently outperform classical model-driven methods (e.g., NGMeet, LRTFDFR, HyMix) on complex, heterogeneous noise, owing to their greater adaptivity and representation capacity.

On the ICVL dataset, DU-TRPCA delivers consistently superior denoising results compared to all baselines. In particular, our method demonstrates clear advantages over both HSDT and its larger-capacity variant HSDT\_L, which serve as critical references since the sparse transformer in DU-TRPCA is based on the GSSA mechanism. Notably, owing to our stage-wise parameter sharing scheme, DU-TRPCA achieves strong denoising performance with only 1.05M parameters-nearly half that of HSDT\_L (2.09M), and comparable to other lightweight baselines (see Table~\ref{tab:gauss_icvl}). This confirms that our performance gains do not simply stem from increased model capacity, but rather from the principled stage-wise alternation of low-rank and sparse priors as motivated by TRPCA. This observation is further substantiated by ablation experiments (see Section~\ref{sec:ablation}). Compared with other state-of-the-art deep unfolding, transformer-based, and hybrid-driven methods, DU-TRPCA achieves consistently better results and displays remarkable robustness to diverse forms of structured and sparse noise. 

To further corroborate these quantitative results, Figure~\ref{fig:icvl_mixture_2rows} presents a visual comparison on the mixture-noise scenario of ICVL's Labtest\_0910-1510. As highlighted in the zoom-in boxes, classical model-driven and early deep learning methods exhibit clear residual artifacts and over-smoothing. Among hybrid methods, those that involve multiple priors but lack stage-wise alternation (e.g. architecture-guided T3SC and deep unfolding-based SMDSNet) produce large-scale deviations and fail to eliminate structured noise. Methods considering only low-rank priors (e.g. architecture-guided HyLora and deep unfolding-based ILRNet) still leave obvious deadline artifacts. In contrast, DU-TRPCA achieves the cleanest restoration, with minimal residual noise and superior structural fidelity, even in regions where other methods show persistent artifacts or blurring. These visual results collectively demonstrate the advantage of an explicit stage-wise alternation between low-rank and sparse priors. Furthermore, Figure~\ref{fig:icvl_mixture_2rows} shows that our method yields reflectance spectra at a representative pixel (161,376) closest to the ground truth, with greatly reduced spectral distortion across all noise bands. This highlights DU-TRPCA’s capacity for both effective mixed noise removal and superior preservation of essential spectral information, which is vital for downstream hyperspectral analysis.

Similar trends are observed on the CAVE dataset, where DU-TRPCA ranks first on all quantitative metrics. Notably, the performance gap widens under the data-limited setting of CAVE, underscoring the strong generalization and data efficiency brought by our physically motivated priors.

Unlike previous works that treat low-rank and sparse modeling independently or combine them heuristically, DU-TRPCA strictly enforces an explicit, interpretable alternation between thresholded t-SVD and Top-K sparse transformer modules. This theoretical rigor translates to robust mutual refinement of priors and consistently superior denoising across noise types. Notably, methods such as ILRNet and HyLoRa, although highly competitive in certain settings, are less robust to complex mixture noise due to the lack of explicit stage-wise sparse modeling.

Collectively, these results establish that the TRPCA-inspired deep unfolding paradigm—characterized by explicit, interpretable alternation between low-rank and sparse priors—offers superior denoising performance, robustness, and generalization under mixed noise conditions. This provides strong evidence for the effectiveness and practicality of physically-motivated, tightly-coupled network designs in hyperspectral image restoration.

\textit{2) Gaussian Noise:}
Tables~\ref{tab:gauss_icvl} and~\ref{tab:gauss_cave} summarize the quantitative denoising results of all competing methods under additive white Gaussian noise with varying standard deviations on the ICVL and CAVE datasets.

For Gaussian noise, leading data-driven approaches—especially those utilizing advanced transformer architectures and deep unfolding frameworks—generally achieve the strongest results. However, it is noteworthy that hybrid models incorporating low-rank priors, such as T3SC, HyLoRa, and ILRNet, also deliver highly competitive performance under Gaussian noise. This advantage becomes especially evident on the CAVE dataset, where the limited number of training samples makes the role of model-driven regularization more prominent.

Our ablation experiments further validate the utility of low-rank priors in this setting: when we introduce a low-rank module into the transformer backbone (see the Ours(-TopK) row), there is a clear improvement in denoising performance for Gaussian noise. This demonstrates that even with powerful transformer-based architectures, explicit low-rank modeling can further enhance denoising, particularly in small-sample regimes.

In contrast, when we additionally incorporate the TopK sparse selection block (Ours row), the performance in Gaussian noise scenarios actually declines. This observation supports our theoretical expectation: for purely Gaussian noise, explicit sparse constraints do not bring additional benefits, and may even negatively impact the model’s performance.

\textit{3) Robustness to Different Noise Types: Gaussian vs. Impulse}
\begin{table}[!htbp]
  \caption{PSNR Robustness Under Noise Type Shift: Gaussian $\rightarrow$ Impulse}
  \label{tab:gauss2imp}
  \centering\footnotesize
  \setlength{\tabcolsep}{4pt}
  \begin{tabular}{l|cc|cc}
    \toprule
    \multirow{2}{*}{Method} &
    \multicolumn{2}{c|}{\textbf{ICVL}} &
    \multicolumn{2}{c}{\textbf{CAVE}}\\
    & $\Delta$PSNR$\;\!\downarrow$ & Rel.\,$\downarrow$ (\%) &
      $\Delta$PSNR$\;\!\downarrow$ & Rel.\,$\downarrow$ (\%) \\
    \midrule
    \textbf{Ours}      & \textbf{-1.00} & \textbf{-2.3} & \textbf{-2.09} & \textbf{-5.1}\\
    HSDT\_L            & -1.77 & -4.1 &  ---  & ---\\
    HSDT               & -2.28 & -5.3 & -3.48 & -8.6\\
    ILRNet             & -2.50 & -5.8 & -2.64 & -6.5\\
    HyLoRa             & -8.60 & -19.8 & -6.13 & -15.4\\
    T3SC               & -5.56 & -13.4 & -6.41 & -16.3\\
    SMDSNet            & -10.02& -26.2 & -9.21 & -25.9\\
    \bottomrule
  \end{tabular}
\end{table}

We further analyze the robustness of each method by comparing its performance under dense, zero-mean Gaussian noise and sparse, high-magnitude impulse noise. This comparison offers insight into how well different architectural designs cope with fundamentally different noise structures.

As summarized in Table~\ref{tab:gauss2imp}, DU-TRPCA achieves the smallest performance drop when moving from Gaussian to impulse noise, with only a 2.3\% decrease in PSNR on ICVL and 5.1\% on CAVE. This highlights the benefit of the explicit, stage-wise alternation between low-rank and sparse priors, enabling DU-TRPCA to maintain strong denoising capability regardless of whether the corruption is dense or sparse.

In contrast, increasing model capacity (e.g., HSDT\_L) or stacking low-rank modules (e.g., ILRNet) can partially alleviate the degradation, but remain less robust than our approach. Methods that lack an explicit sparse component—such as HyLoRa, T3SC, and SMDSNet—show dramatic drops in performance, with relative PSNR decreases as high as 26\%. This further confirms that solely relying on low-rank modeling is insufficient to handle sparse, high-magnitude corruptions.

\textit{4) Generalization and Cross-Dataset Transfer:}
\begin{table}[!htbp]
      \centering
      \caption{Generalization Performance on ICVL: Models Trained on CAVE and Tested on ICVL}
      \begin{tabular}{lcccc}
      \toprule
      Method & PSNR & SSIM & SAM  \\
      \midrule
      QRNN3D          & 33.08 & 0.8342 & 0.1843 \\
      T3SC            & 32.53 & 0.8822 & 0.2477 \\
      SMDSNet         & 26.8  & 0.8007 & 0.3501 \\
      TRQ3D           & 34.64 & 0.9031 & 0.1559 \\
      SST             & 30.52 & 0.8249 & 0.2145 \\
      SERT            & 31.42 & 0.8448 & 0.2006 \\
      HyLora          & 32.47 & 0.8777 & 0.1776 \\
      ILRNet          & \underline{37.81} & \underline{0.945} & \underline{0.1072} \\
      HSDT            & 37.68 & 0.9335 & 0.132 \\
      ours & \textbf{39.34}  & \textbf{0.9572} & \textbf{0.1011} \\
      \bottomrule
      \end{tabular}
      \label{tab:icvl}
\end{table}

To further evaluate the generalization capacity of the proposed DU-TRPCA, we conduct a cross-dataset transfer experiment: all models are trained on the CAVE dataset under the mixture noise setting, and tested directly on the ICVL dataset (Table~\ref{tab:icvl}). This challenging setup requires models to handle substantial domain shifts in both spatial-spectral content and noise characteristics, which is highly relevant to real-world scenarios where annotated hyperspectral data are often scarce or unavailable for the target domain.

As shown in Table~\ref{tab:icvl}, DU-TRPCA achieves the best generalization performance across all metrics, surpassing other state-of-the-art methods in PSNR, SSIM, and SAM. In particular, DU-TRPCA achieves a PSNR of 39.34~dB and an SSIM of 0.9572, outperforming the strong baseline ILRNet by a clear margin. Compared to advanced deep unfolding, transformer-based, and hybrid-driven networks, DU-TRPCA exhibits noticeably higher resilience to domain and noise distribution shifts.

This performance gain can be attributed to the explicit prior-driven alternation between low-rank and sparse priors in DU-TRPCA, which enables the model to disentangle structured noise and underlying signal features in a data-adaptive yet interpretable manner. Unlike methods that rely solely on deep representations or heuristic combinations of priors, our approach inherits the robustness and transferability of model-based paradigms while maintaining the flexibility and expressiveness of deep learning.

\subsection{Results on Real Noise}

\graphicspath{{figures/urban/} }
\begin{figure*}[!htbp] 
  \captionsetup[subfigure]{%
    font=scriptsize,        
    labelfont=bf,           
    skip=1pt,               
    justification=centering 
  }
  \centering
  \foreach \i/\imgfile/\algoname in {
    1/gt/Noisy,
    2/NGmeet/NGmeet,
    3/LRTFDFR/LRTFDFR,
    4/Hymix/HyMix,
    5/qrnn3d/QRNN3D,
    6/smdsnet/SMDSNet,
    7/ilrnet/ILRNet,
    8/hsdt/HSDT,
    9/hsdt_l/HSDT\_L,
    10/ours/Ours
  }{
    \begin{subfigure}[t]{0.15\textwidth}  
      \centering
      \begin{tikzpicture}[inner sep=0,
          spy using outlines={
            rectangle,                 
            magnification=4,           
            size=0.54\linewidth,       
            connect spies={draw=red, thick}, 
            every spy on node/.style={draw=red},  
            every spy in node/.style={draw=red}   
          }]
        \node (img) {\includegraphics[width=\linewidth]{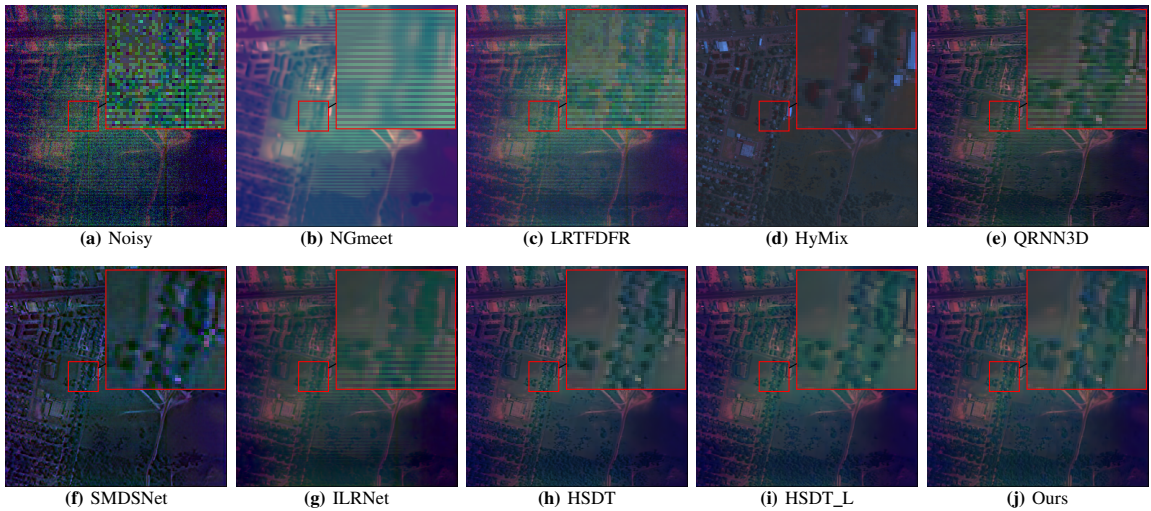}};

        \coordinate (region) at ($(img.center)+(-0.15\linewidth,-0\linewidth)$);

        \spy on (region)
            in node [anchor=north east,
                     xshift= -0.01\linewidth,   
                     yshift= -0.02\linewidth]   
            at (img.north east);
      \end{tikzpicture}
      \caption{\scriptsize \algoname}
      \label{sub:\imgfile}
    \end{subfigure}%
    \ifnum\i=5 \par\medskip\fi
  }
  \caption{Denoising comparison on the real scenario of \texttt{Urban}.}
  \label{fig:urban_2rows}
\end{figure*}

\graphicspath{{figures/indian/}}
\begin{figure*}[!htbp]
  \captionsetup[subfigure]{%
    font=scriptsize,        
    labelfont=bf,           
    skip=1pt,               
    justification=centering 
  }
  \centering
  \foreach \i/\imgfile/\algoname in {
    1/gt/Noisy,
    2/NGmeet/NGmeet,
    3/LRTFDFR/LRTFDFR,
    4/Hymix/HyMix,
    5/qrnn3d/QRNN3D,
    6/smdsnet/SMDSNet,
    7/ilrnet/ILRNet,
    8/hsdt/HSDT,
    9/hsdt_l/HSDT\_L,
    10/ours/Ours
  }{
    \begin{subfigure}[t]{0.15\textwidth}  
      \centering
      \includegraphics[width=\linewidth]{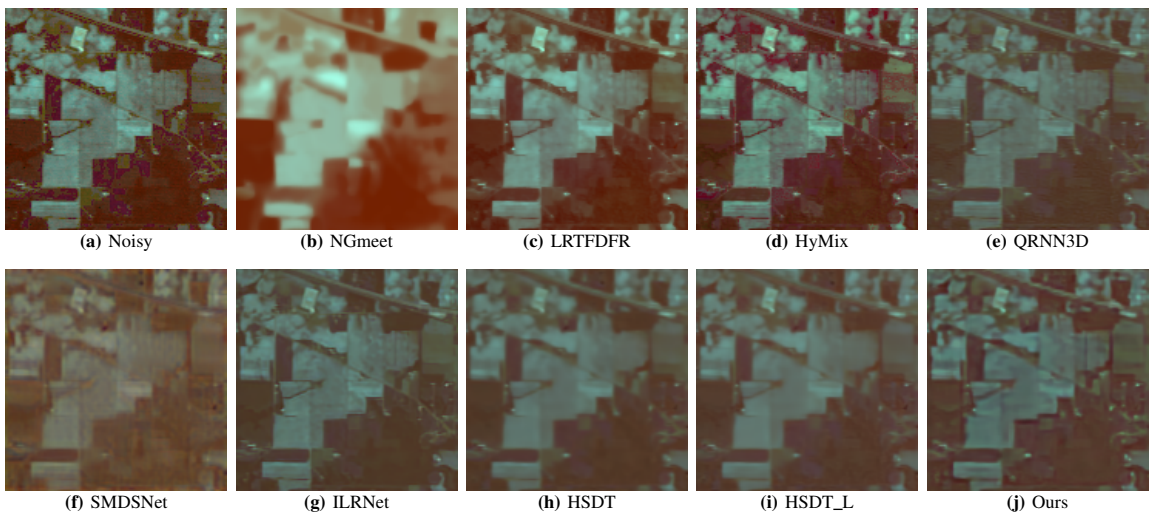}
      \caption{\algoname}
      \label{sub:\imgfile}
    \end{subfigure}%
    \ifnum\i=5 \par\medskip\fi
  }
  \caption{Denoising comparison on the real scenario of 
    \texttt{Indian\_Pines}.}
  \label{fig:indian_2rows}
\end{figure*}

To ensure a fair and practical evaluation on real-world hyperspectral images, we focus on methods that support flexible spectral band configurations—a property essential for deployment in real applications with variable sensors. Among all baselines, only QRNN3D, SMDSNet, ILRNet, and HSDT possess this flexibility; all other methods require retraining to adapt to the real datasets. For this comparison, we directly apply the models trained on ICVL mixture noise to the real Urban and Indian Pines datasets without any fine-tuning.

Figures~\ref{fig:urban_2rows} present qualitative results on the Urban dataset, which exhibits complex, spatially variant, and structured noise. While existing data-driven and hybrid methods can partially suppress such corruptions, our DU-TRPCA achieves notably cleaner visual results and better structure preservation. Notably, despite having a lower parameter count, DU-TRPCA consistently restores sharper and more accurate road boundaries and intersections in challenging regions (see zoomed-in boxes), demonstrating its efficiency and strong real-scene adaptability.

Similarly, on the Indian Pines dataset (Figure~\ref{fig:indian_2rows}), DU-TRPCA delivers the best visual quality, effectively removing noise while preserving spatial boundaries and spectral consistency. Competing methods, by contrast, often leave residual artifacts, over-smooth homogeneous regions, or blur important scene structures.

These real-data results further highlight the practical merits of our TRPCA-inspired approach: by explicitly integrating low-rank and sparse priors within a compact unfolding network, DU-TRPCA offers superior noise removal and detail retention—even under the challenging conditions of real hyperspectral imaging—outperforming both classical and modern learning-based baselines in scenarios requiring spectral flexibility and robust generalization.

\begin{table*}[!htbp]
      \centering
      \caption{Effect of Stage Number on Denoising Performance}
      \resizebox{\textwidth}{!}{%
      \begin{tabular}{l|cccc|cccc|cccc|cccc|cccc}
      \toprule
      \multirow{2}{*}{Method} & \multicolumn{4}{c|}{mix} & \multicolumn{4}{c|}{deadline} & \multicolumn{4}{c|}{impulse} & \multicolumn{4}{c|}{noiid} & \multicolumn{4}{c}{stripe} \\
      \cmidrule(lr){2-5} \cmidrule(lr){6-9} \cmidrule(lr){10-13} \cmidrule(lr){14-17} \cmidrule(lr){18-21}
       & PSNR & SSIM & SAM & Time & PSNR & SSIM & SAM & Time & PSNR & SSIM & SAM & Time & PSNR & SSIM & SAM & Time & PSNR & SSIM & SAM & Time \\
      \midrule
      D2 & 37.741 & 0.929 & 0.229 & 3.156 & 39.900 & 0.948 & 0.158 & 3.148 & 37.893 & 0.925 & 0.253 & 3.143 & 40.186 & 0.951 & 0.156 & 3.129 & 39.568 & 0.945 & 0.173 & 3.136 \\
      D3 & \underline{38.369} & \underline{0.939} & \underline{0.202} & 4.738 & 40.766 & 0.965 & 0.120 & 4.722 & 38.308 & 0.923 & \underline{0.231} & 4.737 & \underline{41.030} & 0.965 & 0.122 & 4.723 & \underline{40.470} & \underline{0.961} & 0.135 & 4.710 \\
      D4 & \textbf{38.759} & \textbf{0.940} & \textbf{0.193} & 6.336 & \textbf{41.005} & \textbf{0.967} & \textbf{0.115} & 6.314 & \textbf{38.674} & \textbf{0.928} & 0.237 & 6.251 & \textbf{41.235} & \textbf{0.968} & \textbf{0.116} & 6.259 & \textbf{40.662} & \textbf{0.964} & \textbf{0.127} & 6.312 \\
      D5 & 37.920 & 0.930 & 0.219 & 7.943 & 40.277 & 0.962 & 0.128 & 7.849 & 38.045 & 0.921 & 0.237 & 7.825 & 40.559 & 0.962 & 0.130 & 7.834 & 39.958 & 0.957 & 0.145 & 7.861 \\
      D6 & 38.279 & 0.938 & 0.203 & 9.386 & \underline{40.792} & \underline{0.966} & \underline{0.119} & 9.411 & \underline{38.355} & \underline{0.928} & \textbf{0.230} & 9.399 & 41.005 & \underline{0.966} & \underline{0.121} & 9.384 & 40.368 & 0.961 & \underline{0.133} & 9.546 \\
      \bottomrule
      \end{tabular}%
      }
      \label{tab:stage}
\end{table*}

\begin{table*}[!htbp]
\caption{Ablation Study on Model Components}
\label{tab:ablation}
\centering\scriptsize
\renewcommand{\arraystretch}{1.1}
\begin{adjustbox}{max width=\textwidth}
\begin{tabular}{
  l||
  S[table-format=2.3] S[table-format=1.3] S[table-format=1.3]|
  S[table-format=2.3] S[table-format=1.3] S[table-format=1.3]|
  S[table-format=2.3] S[table-format=1.3] S[table-format=1.3]|
  S[table-format=2.3] S[table-format=1.3] S[table-format=1.3]|
  S[table-format=2.3] S[table-format=1.3] S[table-format=1.3]
}
\toprule
\multirow{2}{*}{Method}
  & \multicolumn{3}{c|}{non‐iid}
  & \multicolumn{3}{c|}{stripe}
  & \multicolumn{3}{c|}{deadline}
  & \multicolumn{3}{c|}{impulse}
  & \multicolumn{3}{c}{mix} \\
\cmidrule(lr){2-4}
\cmidrule(lr){5-7}
\cmidrule(lr){8-10}
\cmidrule(lr){11-13}
\cmidrule(lr){14-16}
  & {PSNR} & {SSIM} & {SAM}
  & {PSNR} & {SSIM} & {SAM}
  & {PSNR} & {SSIM} & {SAM}
  & {PSNR} & {SSIM} & {SAM}
  & {PSNR} & {SSIM} & {SAM} \\
\midrule
HSDT (ICCV’23)
  & 40.336 & 0.958 & 0.171
  & 39.781 & 0.953 & 0.178
  & 40.036 & 0.958        & 0.170
  & 37.113        & 0.885        & 0.292
  & 37.143        & 0.898        & 0.289 \\

Deep Unfolding
  & 40.930        & 0.964        & 0.123
  & 40.370        & 0.960        & 0.132
  & 40.748        & 0.964        & 0.121
  & 38.007        & 0.912        & 0.268
  & 38.107        & 0.926        & 0.234 \\

Deep Unfolding + TopK
  & 40.879        & 0.966        & 0.121
  & 40.291        & 0.961        & 0.131
  & 40.518        & 0.965 & 0.121
  & 38.060 & {\underline{0.918}} & {\underline{0.239}}
  & 37.994 & {\underline{0.929}} & {\underline{0.215}} \\

Deep Unfolding + t-SVD
  & \textbf{41.617} & \textbf{0.970} & \textbf{0.112}
  & \textbf{41.034} & \textbf{0.966} & \textbf{0.119}
  & \textbf{41.411} & \textbf{0.969} & \textbf{0.111}
  & {\underline{38.396}} & 0.914        & 0.262
  & {\underline{38.527}} & 0.927        & 0.227 \\

Deep Unfolding + t-SVD + TopK
  & {\underline{41.235}} & {\underline{0.968}} & {\underline{0.116}}
  & {\underline{40.662}} & {\underline{0.964}} & {\underline{0.127}}
  & {\underline{41.005}} & {\underline{0.967}} & {\underline{0.115}}
  & \textbf{38.674} & \textbf{0.928} & \textbf{0.237}
  & \textbf{38.759} & \textbf{0.940} & \textbf{0.193} \\

\bottomrule
\end{tabular}
\end{adjustbox}
\end{table*}

\subsection{Ablation Study}\label{sec:ablation}

To comprehensively evaluate the contribution of each module and the overall network design, we conduct detailed ablation studies on the CAVE dataset under various noise types.

\textbf{1) Stage Analysis:}  
Table~\ref{tab:stage} presents the performance of DU-TRPCA with different numbers of unfolding stages (stage2–stage6). As the number of stages increases, denoising performance improves initially but gradually saturates. Notably, the four-stage setting (stage4) achieves the best trade-off between denoising accuracy and computational efficiency, with further increases yielding only marginal gains at the cost of additional inference time. This empirically validates our choice of network depth for all subsequent experiments.

\textbf{2) Module Analysis:}
Table~\ref{tab:ablation} presents a comprehensive module ablation on the CAVE dataset, highlighting the distinct roles and interplay of each architectural component under various noise conditions. Beginning with the baseline HSDT, we successively introduce deep unfolding, Top-K sparse attention, and the thresholded t-SVD low-rank module. The results reveal several important observations. First, while deep unfolding brings moderate and consistent improvements in denoising stability, the effect of individual priors is more nuanced. The Top-K sparse attention module notably enhances robustness against impulsive and outlier-heavy noise, but its contribution is limited—or even slightly negative—under pure Gaussian noise, consistent with the notion that explicit sparsity benefits structured rather than dense corruptions. Conversely, the thresholded t-SVD module offers substantial advantages in handling structured and spectrally correlated noise, as well as in low-sample regimes, but is less effective against localized sparse noise.

Most importantly, it is the explicit, stage-wise alternation of the low-rank and sparse modules—as implemented in DU-TRPCA—that yields the most consistent and pronounced performance gains, especially under challenging mixture and impulse noise. This alternating scheme enables the model to disentangle and suppress both dense and sparse corruptions in a theoretically grounded and interpretable manner, directly reflecting the TRPCA principle.

Overall, these ablation results underscore that while each module addresses specific noise characteristics, only their explicit integration in an alternating framework ensures robust and generalizable denoising across diverse, real-world hyperspectral scenarios, fully validating both the architectural design and theoretical foundation of DU-TRPCA.

\section{Conclusion}
In this work, we presented a TRPCA-inspired deep unfolding network called DU-TRPCA for hyperspectral image denoising, which explicitly alternates between thresholded t-SVD modules for low-rank recovery and Top-K sparse transformer modules for robust outlier suppression. By closely integrating model-based priors with data-driven learning, our approach effectively leverages the strengths of both paradigms while maintaining theoretical interpretability. Extensive experiments on both synthetic and real-world datasets demonstrate that DU-TRPCA achieves state-of-the-art denoising performance, particularly under challenging mixed noise conditions, with high parameter efficiency. Ablation studies further confirm the effectiveness of each component and highlight the critical role of stage-wise alternation. Moreover, DU-TRPCA shows strong generalization across datasets and noise types, underlining its practical value for real-world hyperspectral restoration tasks. Looking forward, we believe that such a tightly coupled prior-driven and data-driven framework offers a promising direction for advancing robust hyperspectral image analysis and restoration in broader applications.


%

\section*{Acknowledgment}

This study was supported by Natural Science Foundation of Shandong Province (No. ZR2021MF104, No. ZR2021MF113), National Natural Science Foundation (No. 62072288), Open Project of National Key Laboratory of Large scale Personalized Customization System and Technology (No.H\&C-MPC-2023-02-04).

\ifCLASSOPTIONcaptionsoff
  \newpage
\fi



\bibliographystyle{IEEEtran}
\bibliography{ref}

%

%

\begin{IEEEbiography}[{\includegraphics[width=1in,height=1.25in,clip,keepaspectratio]{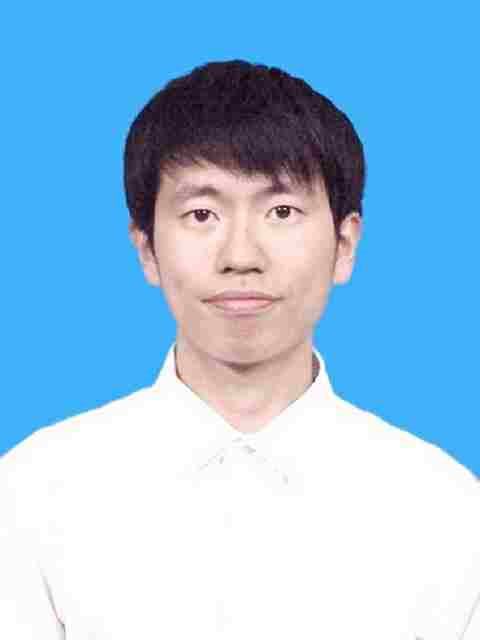}}]{Liang Li}
received his M.E. degree in Computer Science and Technology in 2023 and is currently a Ph.D. candidate at Shandong University of Science and Technology, Qingdao, China. His research interests include image restoration, deep unfolding networks, and tensor decomposition methods.
\end{IEEEbiography}

\begin{IEEEbiography}[{\includegraphics[width=1in,height=1.25in,clip,keepaspectratio]{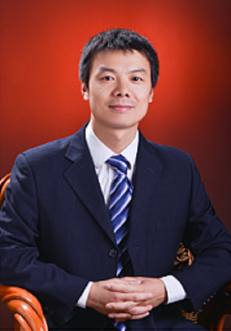}}]{Jian-li Zhao}
received his PhD degree from the Department of Computer Application Technology, Northeastern University, China. He is a Ph.D. supervisor and currently serves as Associate Dean of the College of Computer Science and Engineering at Shandong University of Science and Technology. He is a Distinguished Member of the China Computer Federation (CCF), Chair of the CCF Qingdao Chapter, recognized as a “Haiyou Innovation Talent” in Jinan, and a Top Talent in Qingdao West Coast New Area. His main research interests include tensor decomposition, image restoration, and recommender systems.
\end{IEEEbiography}

\begin{IEEEbiography}[{\includegraphics[width=1in,height=1.25in,clip,keepaspectratio]{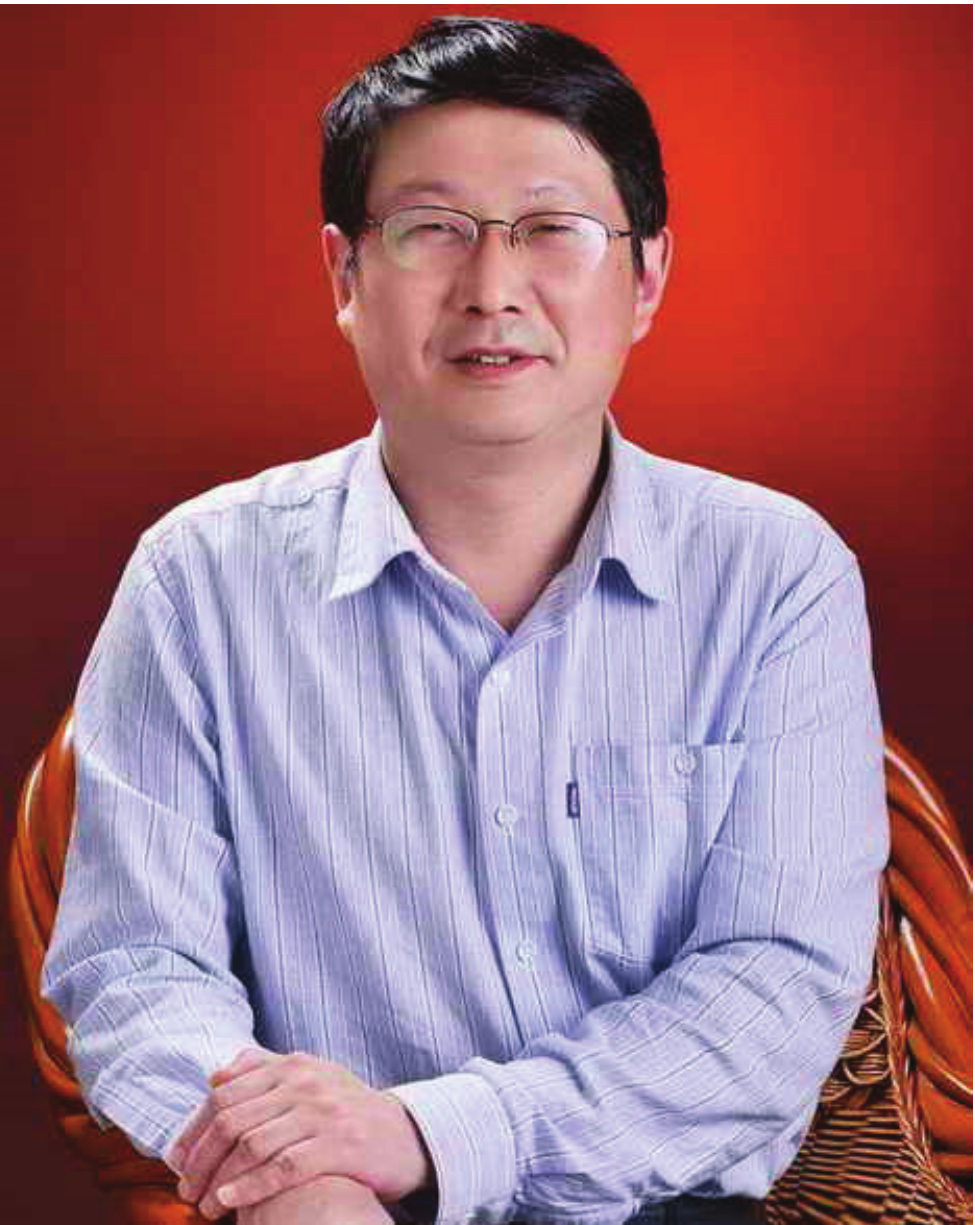}}]{Sheng Fang}
received both the B.S. in Applied Mathematics and B.E. in Material Science from Shanghai Jiaotong University in 1994, M.S. and Ph.D. degrees from Tsinghua University and Shandong University of Science and Technology in Computer Science in 2002 and 2008 respectively.

He is currently a Professor in College of Computer Science and Engineering, Shandong University of Science and Technology. He is a senior member of China Computer Federation, a member of CCF Multimedia Technical Committee, and ACM SIGMM China Chapter Member. His research interests are in the fields of remote sensing image processing, change detection, and object detection. He has published more than 30 papers.
\end{IEEEbiography}

\begin{IEEEbiography}[{\includegraphics[width=1in,height=1.25in,clip,keepaspectratio]{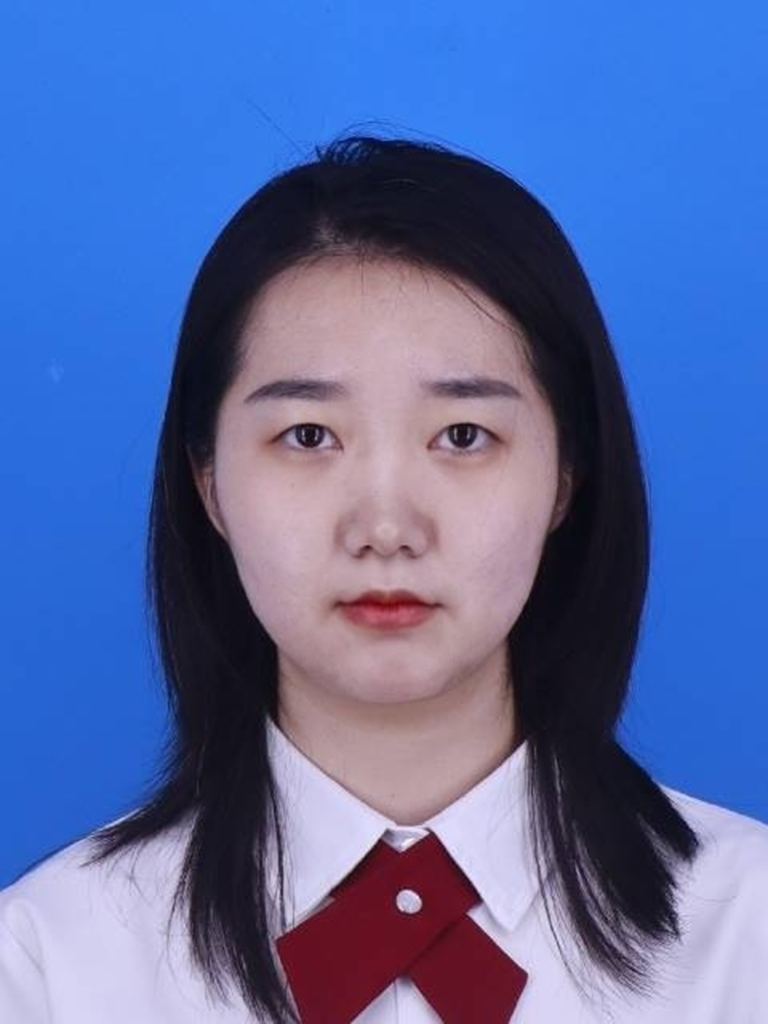}}]{Si-yu Chen}
received the M.E. degree in 2023 form Ludong University. She is currently pursuing the Ph.D. degree with the School of Computer Science and Engineering, Shandong University of Science and Technology, Qingdao, China. Her research interests include brain-like model optimization, such as model compression and redundancy removal.
\end{IEEEbiography}

\begin{IEEEbiography}[{\includegraphics[width=1in,height=1.25in,clip,keepaspectratio]{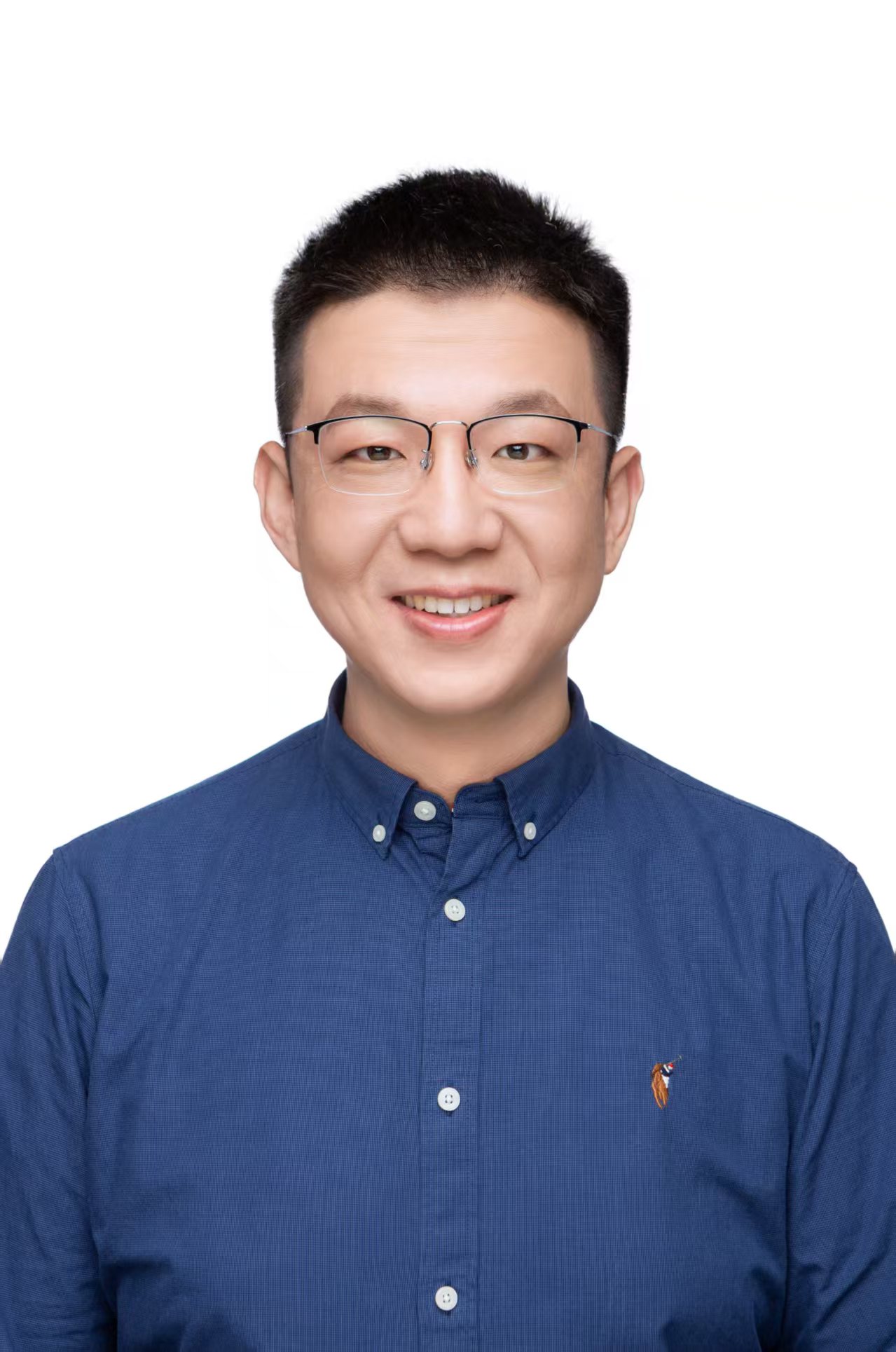}}]{Hui Sun}
received his M.E. degree in Microelectronics and Solid-State Electronics from Nankai University. He is currently a senior executive at Gosci Information Technology Co.,Ltd. His research interests include marine data processing, big data, and artificial intelligence.
\end{IEEEbiography}




\end{document}